\documentclass[utf8]{frontiersSCNS} 
\usepackage{url,hyperref,lineno,microtype,subcaption}

\usepackage{filecontents}
\usepackage{caption}
\usepackage{subcaption}
\usepackage{amsmath}
\usepackage{mattens}
\usepackage{siunitx}
\usepackage{booktabs}
\usepackage{graphicx} 
\usepackage{lipsum,lineno}
\usepackage{pgfplots}
\usepackage{pgfplotstable}
\usepgfplotslibrary{groupplots}
\usepackage{threeparttable}

\pgfplotsset{compat=1.13}
\usepackage{amsmath}
\usepackage{tikz} 
\usepackage{tikz-3dplot} 
\pgfplotsset{compat=newest}
\usetikzlibrary{decorations.markings}
\usetikzlibrary{spy}
\usepackage{amssymb,amsfonts,amsmath}
\usetikzlibrary{arrows,calc,patterns}
\usetikzlibrary{math}
\usepackage{mathtools, cuted}
\usepackage{newtxtext,newtxmath}
\usepackage{bm}

\DeclareRobustCommand{\uvec}[1]{{%
  \ifcsname uvec#1\endcsname
     \csname uvec#1\endcsname
   \else
    \bm{\hat{\mathbf{#1}}}%
   \fi
}}
\definecolor{gold}{RGB}{212,187,106}
\definecolor{bluish}{RGB}{18,54,82}
\definecolor{orange}{RGB}{243,114,32}

\pgfplotsset{select coords between index/.style 2 args={
        x filter/.code={
            \ifnum\coordindex<#1\fi
            \ifnum\coordindex>#2\fi
                              }
    }
}

\tikzset{new spy style/.style={spy scope={
     magnification=5,
     size=1.25cm, 
     connect spies,
     every spy on node/.style={rectangle,draw},
     every spy in node/.style={draw,rectangle}
      }
   }
} 

\usetikzlibrary{decorations.markings}
\usepackage{scalerel}
\usepackage{tikz}
\usetikzlibrary{svg.path}
\usepgfplotslibrary{colorbrewer}
\definecolor{orcidlogocol}{HTML}{A6CE39}
\tikzset{
  orcidlogo/.pic={
    \fill[orcidlogocol] svg{M256,128c0,70.7-57.3,128-128,128C57.3,256,0,198.7,0,128C0,57.3,57.3,0,128,0C198.7,0,256,57.3,256,128z};
    \fill[white] svg{M86.3,186.2H70.9V79.1h15.4v48.4V186.2z}
                 svg{M108.9,79.1h41.6c39.6,0,57,28.3,57,53.6c0,27.5-21.5,53.6-56.8,53.6h-41.8V79.1z M124.3,172.4h24.5c34.9,0,42.9-26.5,42.9-39.7c0-21.5-13.7-39.7-43.7-39.7h-23.7V172.4z}
                 svg{M88.7,56.8c0,5.5-4.5,10.1-10.1,10.1c-5.6,0-10.1-4.6-10.1-10.1c0-5.6,4.5-10.1,10.1-10.1C84.2,46.7,88.7,51.3,88.7,56.8z};
  }
}

\newcommand\orcidicon[1]{\href{https://orcid.org/#1}{\mbox{\scalerel*{
\begin{tikzpicture}[yscale=-1,transform shape]
\pic{orcidlogo};
\end{tikzpicture}
}{|}}}}

\usepackage{hyperref} 
\usepackage{xcolor}

\def\keyFont{\fontsize{8}{11}\helveticabold }
\def\firstAuthorLast{Alejandro Macario-Rojas {et~al.}} 
\def\Authors{Alejandro Macario-Rojas \,$^{1,*}$, Ben Parslew\,$^{1,2}$, Andrew Weightman\,$^{1}$, and Katharine L. Smith\,$^{1}$}

\def\Address{$^{1}$Department of Mechanical, Aerospace and Civil Engineering, The University of Manchester, Manchester M13 9PL, United Kingdom \\
$^{2}$International School of Engineering, Faculty of Engineering, Chulalongkorn University, Bangkok 10330, Thailand  }

\def\corrAuthor{Corresponding Author}
\def\corrEmail{alejandro.macariorojas@manchester.ac.uk}

\begin{document}
\onecolumn
\firstpage{1}

\title[A Minimally Actuated Jumping Robotic Platform]{CLOVER Robot: A Minimally Actuated Jumping Robotic Platform for Space Exploration} 

\author[\firstAuthorLast ]{\Authors} 
\address{} 
\correspondance{} 

\extraAuth{}

\maketitle

\begin{abstract}

\section{}
Robots have been critical instruments to space exploration by providing access to environments beyond human limitations. Jumping robot concepts are attractive solutions to negotiate complex terrain. However, among the engineering challenges to overcome to enable jumping robot concepts for sustained operation, reduction of mechanical failure modes is one of the most fundamental. 
 {\textcolor{black}{This study set out to develop a jumping robot with focus on minimal actuation for reduced mechanism maintenance. We present the synthesis of a Sarrus-style linkage to constraint the system to a single translational degree of freedom without the use of typical synchronising gears. We delimit the present research to vertical solid jumps to assess the performance of the fundamental main-drive linkage. A laboratory demonstrator assists the transfer of theoretical concepts and approaches.}}
{\textcolor{black}{The laboratory demonstrator performs jumps with 63\% potential-to-kinetic energy conversion efficiency, with a theoretical maximum of 73\%. Satisfactory operation opens up design optimisation and directional jump capability towards the development of a jumping robotic platform for space exploration. 
}}

\tiny
 \keyFont{ \section{Keywords:} Jumping, space, robotics, Sarrus-like, overconstrained, mechanism, design} 
\end{abstract}

\section{Introduction}

The practicality of a mobile robot is often limited by its ability to negotiate obstacles. Various land locomotion methods enabled by limbs, wheels, tracks, body articulation, and non-contact locomotion, are used in mobile robots  to negotiate terrain. Insurmountable obstacles are generally dealt with by a heading change, shortening the mean free path. This results in longer commutes and an increased requirement for steering capability. Conversely, if a robot can overcome all obstacles in its path, then no steering is required and there is potential to reduce the commute distance and time.

\noindent {\textcolor{black}{Walking and crawling locomotion offer high terrain adaptability and obstacle negotiation at the cost of high control complexity; this cost is sometimes warranted in some specific applications, e.g.,~\cite{chignoli2021humanoid,liljeback2010controllability}, for example in vertical climbing of snake-like robots. Wheeled and tracked rovers~\cite{wakabayashi2009design} offer robust mobility whenever sufficient traction between the terrain adaptation system, e.g., cleated wheel, and the terrain exists. The maximum obstacle traversable height in wheeled rovers is closely related to the wheel diameter~(\cite{maimone2006surface}). For example, for the typical rocker-bogie chassis-based planetary rover, the maximum traversable height is roughly 50\% the wheel diameter~(\cite{seeni2010robot}). Whereas jumping robots have been developed that can traverse obstacles several times their own height, such as the $45\si{\kilogram}$  $\sim 0.5\si{\metre}$ height PrOP-F mobile apparatus that was expected to jump $\sim 50\si{\metre}$ on the Mars' moon Phobos~(\cite{harvey2007russian,seeni2010robot,ulamec2011hopper}).}}

\noindent A jumping robot gains kinetic energy through contact reaction force from the ground. If slippage is limited during the contact phase a jumping system has the potential for reduced frictional cost of transport compared to walking or rolling systems. This may also imply reduced design complexity in the terrain adaptation system as contact wearing is reduced in comparison to other means of locomotion. One of the most appealing characteristics of jumping locomotion is the ability to traverse challenging terrain using a reduced number of drivable degrees of freedom. This has the potential to reduce robotic cognitive resources and sustained power consumption during locomotion.

\noindent {\textcolor{black}{Jumping locomotion has gained great attention due to the aforementioned characteristics, which may enable small rover concepts with substantial mean free paths and modest control requirements. This has led to several jumping robots being designed for planetary~(\cite{plecnik2017design,zhang2020biologically}), and asteroidal exploration~(\cite{seeni2010robot,ulamec2011hopper,ellery2015planetary,watanabe2017hayabusa2}). However, there are significant unmet needs in jumping exploration robot technologies. For instance, practical jumping robots for space exploration have been historically limited to microgravity conditions that have permitted the use of rotational torquers or whiskers to gain satisfactory jump velocity coining the term of mobile landers, e.g., The Mobile Asteroid Surface Scout~(\cite{ho2017mascot}) and the Micro Nano-experimental Robot Vehicle for Asteroids (MINERVA-II) onboard the Hayabusa-2 mission to the Ryugu asteroid~(\cite{watanabe2017hayabusa2}).}} The success in transitioning from current mobile landers to jumping robotic explorers able to operate in more demanding planetary gravitations, such as lunar gravitation, will reside in resilient designs able to cope with unforeseen situations, particularly in situations where direct human control is not possible or greater autonomy is required.

{\textcolor{black}{In this work is reported the development of}} a main-drive linkage platform for planetary jumping robots addressing the reduction of mechanical mobility failures in favour of sustained operation and mission viability. Complementary thrust force vectoring for jump directionality~(\cite{parslew2018dynamics}) is subject of a subsequent work.

\section{Design Goals}
\label{sec:designgoals}

\noindent The main motivation for this work is to establish the mechanical foundation for the development of a nimble robotic platform capable of complex terrain exploration. Space exploration is a quintessential example of such activities. In general, such a robotic platform would be mainly used to access environments beyond human capabilities, and beyond the capabilities of traditional exploration robots e.g., wheeled explorers. To achieve this aim, this work reports the development process of a mechanism with sufficient reliability and versatility as to set the basis of a generic robotic explorer platform. Thus, we delimit our design using the following three constraints.

\noindent 
{\textcolor{black}{Firstly, our design approach is driven by minimally actuated jumping robot concept. Research efforts are directed toward the development of the main-drive linkage responsible of transforming potential energy into jump kinetic energy. Jump directionality poses unique challenges, as discussed by \cite{parslew2018dynamics}, which are out of the scope of the current work. In the context of this work, we define jumping as a single action to propel a body to become airborne, while hopping involves repetitive consecutive jumps. In a real autonomous robot operation, the ability to come to a halt between jumps may be critical in prevision of system state checks, permit eventual system recovery, and feedforward mission task planning, among others. This ability becomes even more relevant in complex or unknown terrain operations. Although the duration of the halt stage may vary significantly for various applications, this condition necessarily discards periodic hopping capability for this design. The rationale behind the selection of minimal actuation is that driving multiple degrees of freedom not only consumes additional onboard power, a paramount concern in robotic explorers, but also adds control and mechanical complexity to the locomotion system. Specifically, in a jumping robot where dynamics evolve rapidly to gain kinetic energy, increased mechanical complexity inherently increases the risk of most fundamental mechanical failure modes. Furthermore, increased system complexity may be detrimental for small and light robotic explorer designs because size and mass are compromised.
}}

\noindent The second design constraint is the objective physical scale of the platform. Over the last five to ten years, there has been an upward trend in investment and demand in robotics for planetary and microgravity applications. This trend is driven by the increasing need for on-orbit satellite servicing, debris removal, on-orbit manufacturing and assembly, and space exploration~(\cite{smith2020artemis,esalavatube}), fostered primarily by accelerated technological advancements in autonomous systems around the world. Additionally, miniaturised satellite platforms for scientific and economic return, such as CubeSats, have boosted low-cost access to space. These conditions suggest the adequacy of the development of a miniaturised robotic platform. Inspired by the CubeSat standard, a sub-kilogram and sub-metre scale for the {\textcolor{black}{demonstrator is set.}}

\noindent The third design constraint is set by the objective operational conditions. Terrain and environment characteristics are of paramount importance, not only regarding navigability, but also for robot endurance. Hostile ambient conditions such as thermal extremes and dust environments may represent sources of mechanical failure. Dust in particular is a common source of operative deterioration affecting movable joints due to its pervasive presence in most environments. Furthermore, hazardous dusts~(\cite{bayliss2003nuclear}) such as those from asbestos, beryllium oxide, and radioactive materials, may transform exploration robots into unwilting dissemination devices. In planetary rovers, robot-soil interaction is exacerbated by the dominant presence of loose regolith,~(\cite{ellery2005environment}), for example lunar regolith is exceptionally adhesive to any surface due to its frictional and electrostatic properties. In all, the above identifies the need of a design approach aiming at reducing or prescinding human mediation in routine robot maintenance processes. In turn such design would contribute to resilient and reliable locomotion.

\noindent The robotic platform design resulting from these design goals is hereinafter referred to as Controlled Leap Operation for a Versatile Exploration Robot CLOVER. The design programme reported in this paper focuses on the take-off development stage, and is outlined by the following main design goals:

\begin{enumerate}
\item
Design for minimal actuation of main-drive jump linkage (no thrust force vectoring)
\item
Focus on linkage mechanism potential-kinetic energy conversion efficiency
\item
Design for reduced linkage mechanism maintenance 
\end{enumerate}

In the following Section~\ref{sec:mechdesign} the theoretical design of the CLOVER is presented.

\section{Mechanism Design}
\label{sec:mechdesign}

\begin{figure} 
\centerline{\includegraphics[width=5.2in]{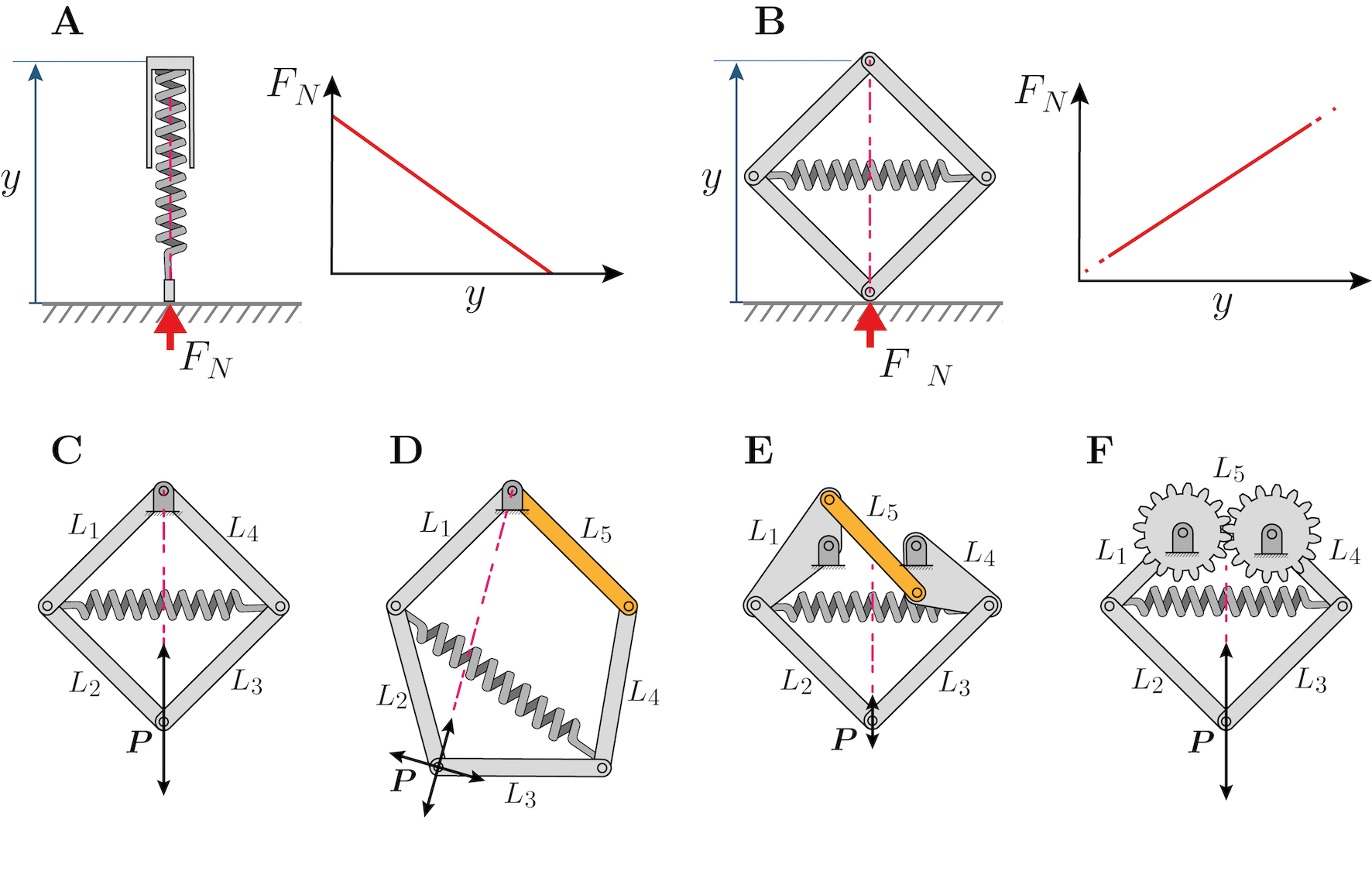}}
\caption{The telescopic prismatic mechanism (A) and hinge rhomb linkage (B), with their characteristic ground reaction force with longitudinal displacement graph. From C--F, synthesis of a synchronous hinge rhomb linkage using a pair of synchronising gears. Dotted line extremes in graph B highlight singular or nonlinear behaviour of $F_N$ with $y$ for specific geometry configurations; these aspects will be discussed in subsequent sections.}
\label{fig_forceacds}
\end{figure}

In line with the first design goal of maintaining minimal actuation, and acknowledging that the jumping process involves mainly the development of linear momentum,  mechanisms with one translational Degree Of Freedom (DOF) are considered. The simplest typical mechanisms that fulfil these criteria include the telescopic prismatic mechanism and the hinge rhomb linkage (Fig.~\ref{fig_forceacds}A and~\ref{fig_forceacds}B respectively). When used within a jumping robot these mechanisms are often spring driven~(\cite{batts2016untethered,carpi2011dielectric,zhao2013msu,hale2000minimally,truong2019design,jung2016integrated}); an electric actuator is used to compress the spring and the actuation force is then removed (e.g., via a mechanical catch) to initiate the take-off phase. Thrust is generated as the ground reaction force acting on the foot of the robot, which accelerates the robot with respect to the ground. In the telescopic prismatic mechanism, the thrust force is maximum when the spring is fully compressed, and reduces linearly with the spring displacement. From a conceptual perspective, this mechanism offers a simple solution to the jumping problem. However, experimental evidence has suggested that this type of mechanism could suffer from spring surge potentially reducing energy conversion efficiency as reported by~\cite{hale2000minimally}.

\noindent An alternative mechanism commonly used for jumping robots is a hinge rhomb linkage~(\cite{zhao2013msu,hale2000minimally,truong2019design,jung2016integrated}). This consists of four links connected by hinge joints (forming two symmetric dyads), and a spring connecting the intermediate joints of the two dyads as shown in Fig.~\ref{fig_forceacds}B. This mechanism has more components inherently adding more complexity to the system in comparison to the telescopic prismatic mechanism\footnote{In order to prevent buckling of the helical compression spring, auxiliary elements are typically used, e.g., guide tubes, rods. However, here these elements are assumed simpler than in the hinge rhomb linkage.} in Fig.~\ref{fig_forceacds}A. However, the mechanism has been cited as a means of removing undesirable spring surge and premature take-offs~(\cite{hale2000minimally}). Also, the linkage acts as an inverter for the thrust force, which implies that maximum thrust occurs at the end and not at the beginning of the jump evolution as occurs in the telescopic prismatic jumping mechanism. This phenomenon reduces the force required to hold the mechanism when at maximum compression, in comparison to the prismatic mechanism. The advantage of this is that the release mechanism (e.g., a latch) would require lower actuation force/torque, and hence lower mass, in the hinge rhomb linkage system than in the prismatic linkage system.

\noindent In the hinge rhomb system, the spring carries tensile loading rather than compression loading. Higher design factors are used in extension springs as these are prone to hooks fatigue and more catastrophic failure than compression springs, representing more mass addition to this system. Notwithstanding, the progressive thrust force and absence of spring surge has shown more practical advantages than disadvantages~(\cite{hale2000minimally}). Furthermore, as the hinge rhomb linkage uses tensile loading, the spring drive can easily be replaced with a flexible material drive with a high specific energy, which allows for alternative designs that may favour reduced weight. From this qualitative analysis, {\textcolor{black}{further synthesis of the proposed design is based on the hinge rhomb linkage.}}

\noindent The implementation of the hinge rhomb linkage in a jumping robot requires the addition of links for functional purposes, e.g., a foot or platforms for onboard payloads. This practical requirement is often fulfilled with the addition of extra links separating the two main dyads acting as legs, i.e., from four links to five or six links. The resulting modified linkage changes the DOF of the end effector ($P$ from Fig~\ref{fig_forceacds}C to D) requiring extra components to recover the original single translational DOF. A common solution to this employs a pair of synchronising gears~(\cite{bai2019design,hale2000minimally}) for the dyads as shown in Fig~\ref{fig_forceacds}F. Planar mechanisms like the one described above are often seen in jumping robot designs. However, this approach is discordant with our third design goal, i.e., design for reduced mechanism maintenance, because exogenous materials can easily compromise operation and reliability of gears and therefore of the whole mechanism. For example, dust or other contaminants can affect tolerances in gears. This in turn affects the dynamic factor, an exponential function of a quality number used to account for inaccuracies in teeth in stress equations~(\cite{budynas2005shigley}); a small variation in the quality number yields large variation in the dynamic factor and therefore in the teeth stress prompting mechanical failure. In contrast contamination tolerance and dust sealing technologies for hinge construction are mature and considerably simpler than those for gears isolation. Bearing technologies have been proven in planetary rovers showing high reliability in extremely challenging and contaminated operation conditions~(\cite{ellery2015planetary}). A contribution of this work is to achieve dyad synchronisation without the use of gears, or any other additional mechanism, by transforming the planar 6-link mechanism into a spatial 6-link mechanism. This can be achieved by creating a new plane through the relative rotation of one dyad with respect to the axis of symmetry defined by the translational vector of the DOF.

\noindent The well-known classical Sarrus mechanism is created when the dyad planes are orthogonal. The Sarrus mechanism is a spatial parallel mechanism with one translational DOF, consisting of two platforms interconnected with two dyads, each of them having three revolute joints ($3-$RRR) perpendicular to the translation axis. Note that despite the elegant simplicity of this alternative to the gear-synchronised 6-link planar mechanism, the line of action of the driver that connects the dyad centres lies outside the plane of the kinematic chain locates (mobility plane); this introduces lateral torques on the revolute joints. A natural way to overcome this new problem is by recovering mechanism symmetry about the axis of translation while preserving the dyads in different planes as mentioned before. This analysis in presented in the following subsection~\ref{kinchainm}.

\hspace{1cm}
\subsection{Kinematic chains and mobility analysis}
\label{kinchainm}

\begin{figure} 
\centerline{\includegraphics[width=2.5in]{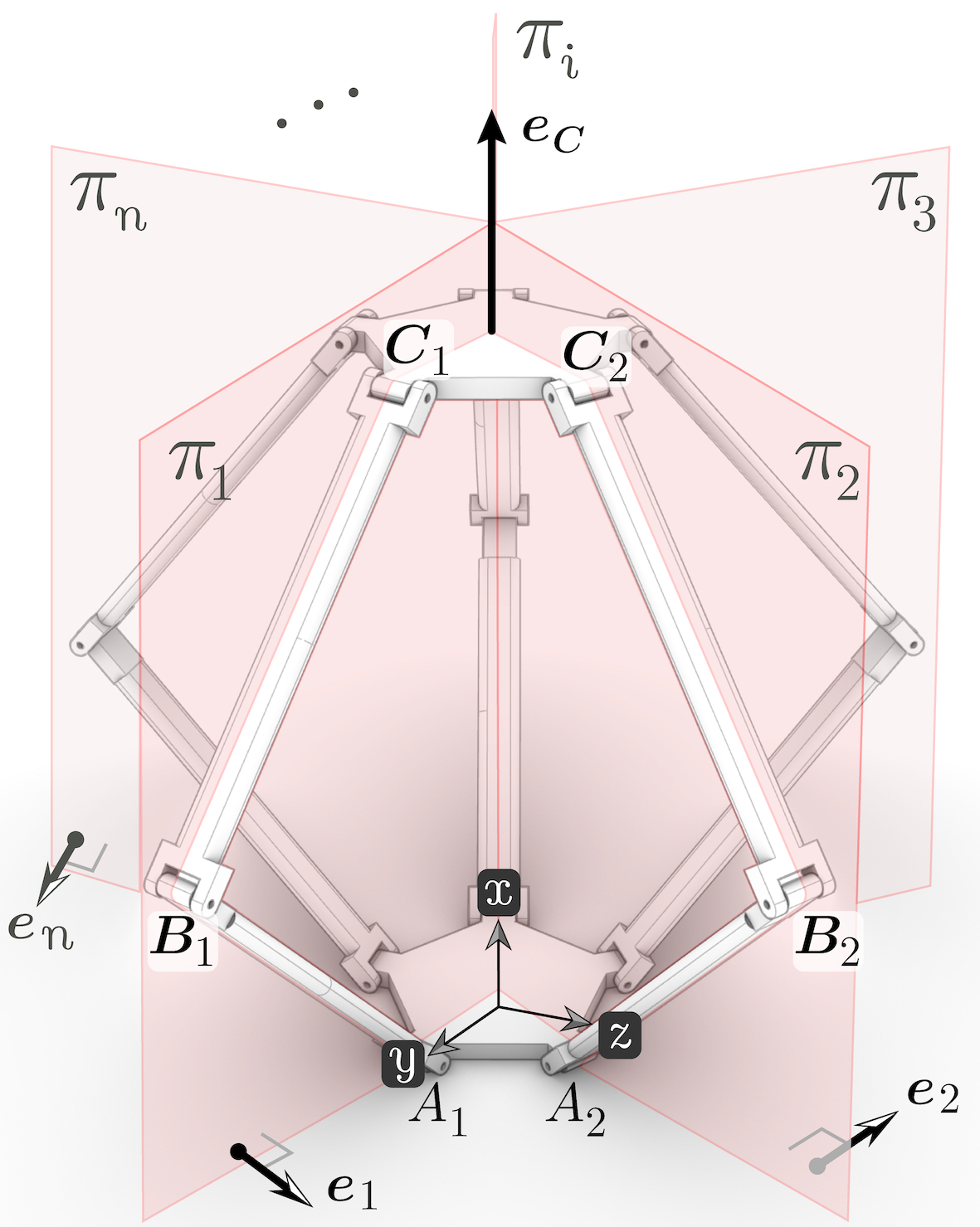}}
\caption{$n-$sided Sarrus linkage}
\label{fig_screwsgeom}
\end{figure}

\noindent In first instance, the mobility of the Sarrus mechanism can be investigated by exploring the geometry characteristics of the mechanism. In a general architecture of a Sarrus-based mechanism in Fig.~\ref{fig_screwsgeom}, an upper and base platform are connected through $n-$RRR kinematic chains identical in topology. Each kinematic chain $i-$RRR with $i=1,2,3, \dots n$ always remains in its mobility plane $\pi_i$. The $i-$th kinematic chain \textit{only} allows its terminal connector, i.e., the one connected to the upper platform in this analysis, independent planar motions on $\pi_i$. Due to the common connection between the $n-$RRR kinematic chains through the upper platform, its mobility is constrained to one translation (up and down) along the common intersecting line of the $\pi_n$ planes. As this line lies in the central line of the mechanism, this general architecture can be defined as a centralised moving parallel mechanism. {\textcolor{black}{From these observations, the mobility analysis of the mechanism can be formalised as follows.}}

\begin{figure} 
\centerline{\includegraphics[width=3.5in]{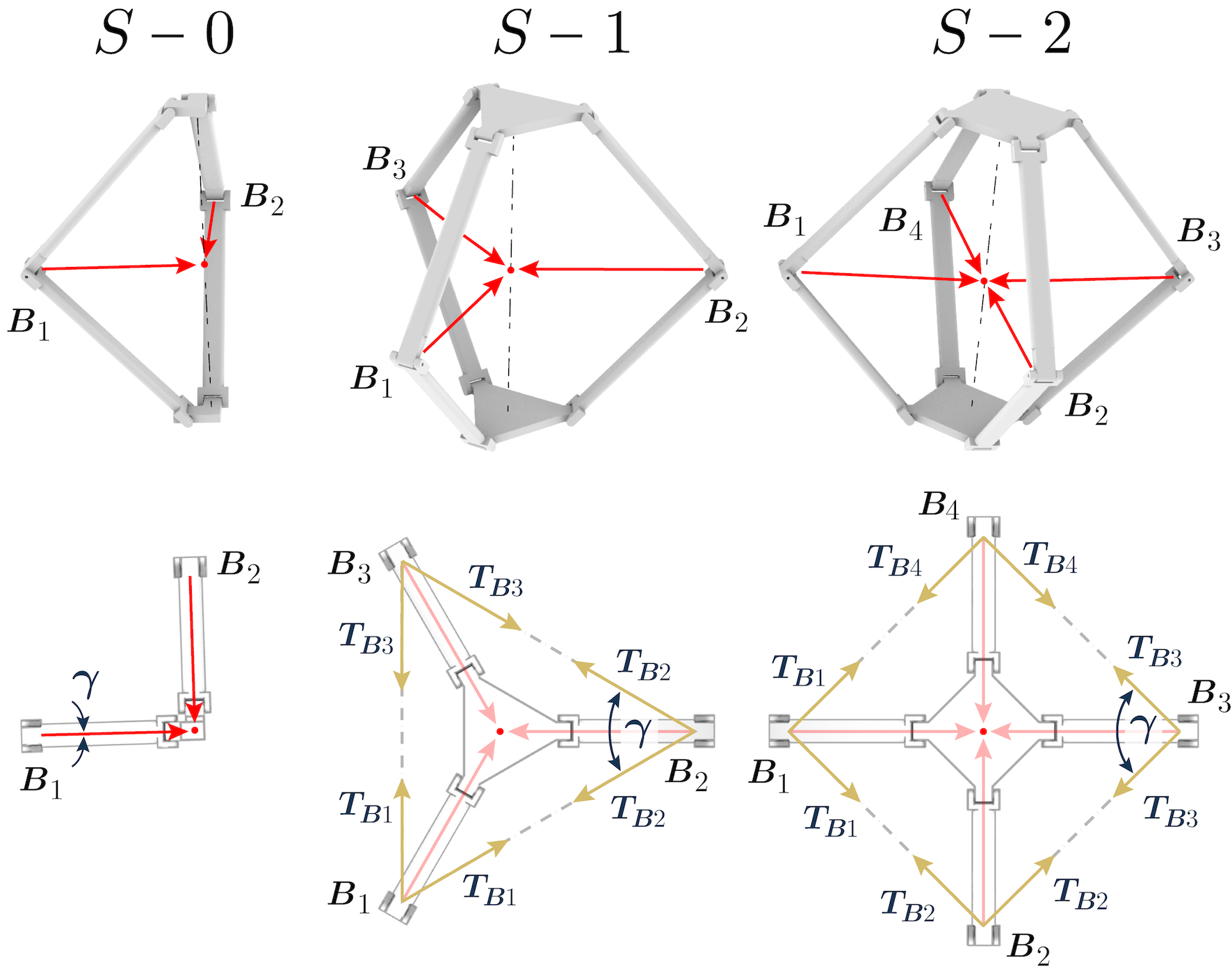}}
\caption{Family of mechanisms and tensile loadings distribution. From left to right, the classical 6-link Sarrus mechanism $S-0$, first dyad redundancy $S-1$, and second dyad redundancy $S-2$, from $S-n$ possible configurations. The arrows show force vectors generating null net torque on the revolute joints.}
\label{fig_fammech_tens}
\end{figure}

\noindent Mobility analysis of mechanisms is typically carried out using the Gr\"{u}bler-Kutzbach Criterion (G-K), which is based on simple arithmetics. The G-K criterion can be successfully applied to almost all planar and some spatial mechanisms. However, when redundant constraint or over-constraint appears in a mechanism G-K fails in most cases~(\cite{huang2013mobility}). For this reason, Screw theory is often used in the analysis of spatial mechanisms. {\textcolor{black}{Appendix~\ref{appendixA:SCT} presents}} the mobility analysis of the over-constrained Sarrus linkage based on screw theory. From that analysis is concluded that the sole condition to preserve one translational DOF in this type of mechanism is that at least two dyads remain in nonparallel planes. Therefore, a family of mechanisms of this kind {\textcolor{black}{can be named}} with respect to a fundamental configuration ensuring the essential mechanical behaviour of the linkage, e.g., the classical Sarrus mechanism as shown in Fig.~\ref{fig_fammech_tens}.

\noindent The presented family of mechanisms also shows tensile loading arrows created by the spring, which originally connected the intermediate joints of the two dyads (hereinafter referred to as knee) in the hinge rhomb linkage in Fig.~\ref{fig_forceacds}A. Note that the first fundamental configuration, $S-0$, shows no rotational symmetry; that is, the angle between the two dyad planes ($n=2$) is different from $2 \pi/n$ radians. Conversely to $S-0$, configurations with rotational symmetry are advantageous from the point of view of load distribution (Fig.~\ref{fig_fammech_tens}). The second configuration $S-1$, highlights one-dyad redundancy with respect to the fundamental configuration $S-0$, with rotational symmetry of order 3 ($120^{\circ}$ between planes). Similarly, other symmetric configurations of higher order can be formed without affecting the mechanical behaviour of the linkage.

\noindent Fig.~\ref{fig_fammech_tens} shows the $S-1$ configuration with tensile loadings $T_B$ at aperture angles $\gamma/2$. Each $T_B$ has two orthogonal components, one projected along the dyad plane that is responsible of actuating the mechanism, and a complementary orthogonal projection adding torque to the root revolute joints. As long as the loads show bilateral symmetry with respect to the link pane, they will produce no net torque on the joints. Notwithstanding, any value of $\gamma\neq0$ prompts stretch reduction, affecting the elastic potential energy of the drive, which is a function of stretch squared.

\subsection{Thrust force}
\label{sec:thf}

\begin{figure} 
\centerline{\includegraphics[width=4.0in]{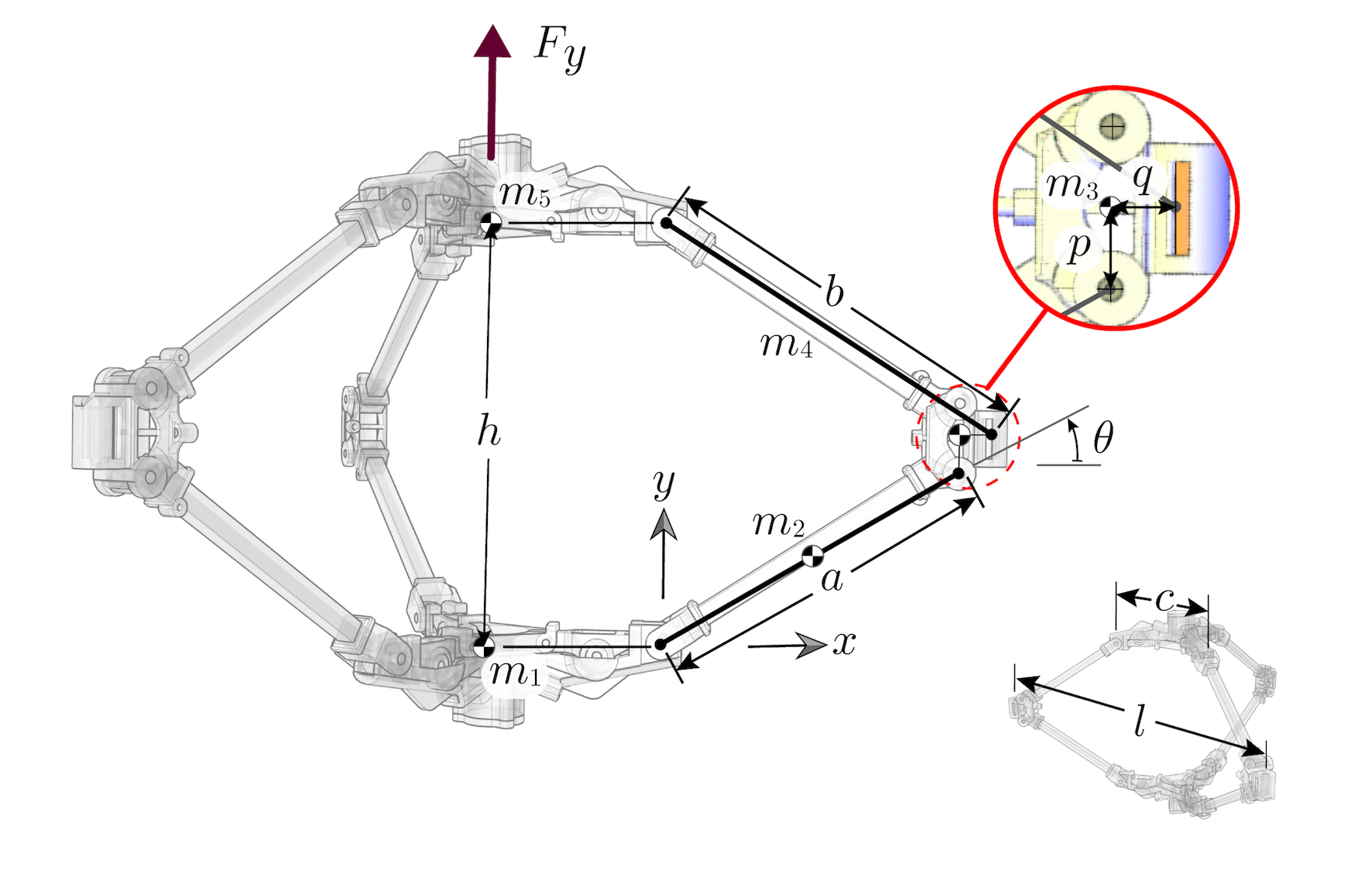}}
\caption{Linkage cage. Elastic elements, not shown in the illustration, are anchored to the knees and run along the distance $l$.}
\label{fig_cad}
\end{figure}

\noindent The approach implemented in the CLOVER robot design is based on the  $S-1$ configuration illustrated in Fig.~\ref{fig_cad}, due to geometric simplicity in line with our third design goal in Section \ref{sec:designgoals}. Additionally, {\textcolor{black}{$\gamma=60^{\circ}$ is set for the sake of a simplified anchorage design for the drive as discussed later in this section.}} The leg length is $a$, leg swing angle is $\theta$, the fixed leg separation distance is $c$, $l$ is the variable distance between fixed points in the knees serving as anchorage points for the drive, the variable linkage height is $h$, and the thrust force is $F_y$. For the sake of practical implementation, assume $b$ as the variable length running from the analysed body-leg joint to the effective anchorage point of the elastic element; $p$ and $q$ are lengths that relate $a$ and $b$. Note that $p$ and $q$ could be used to capture in-plane offsets between the axis of rotation of the knee revolute joint, and the practical anchorage point of the elastic element. In our knee design, $p$ and $q$ are constant lengths capturing manufacturing trade-offs discussed later in Section \ref{sec:proto}.

 \noindent The masses of the components are labelled as $m_i$ from bottom to top, and are used in the following subsection. The value of $h$ can be described with Eq.~\ref{var_h} in terms of the constants $a$, $p$, and the time varying $\theta$.

\begin{equation}
h =  2 a \sin \theta + 2 p
\label{var_h}
\end{equation}

\noindent Similarly, $b$ is given by Eq.~\ref{var_leg}.

\begin{equation}
b^2 =  a^2+p^2+q^2+2 a \left(p \sin \theta + q \cos \theta \right)
\label{var_leg}
\end{equation}

\noindent The distance $l$ is given by Eq.~\ref{eq_elonga}.

\begin{equation}
l =  c  + \frac{\sqrt{12 b^2 h^4 - 3 h^6 }}{2 h^2}
\label{eq_elonga}
\end{equation}

\noindent The thrust force is obtained from the spring force along $l$, $F_{l}$, by applying the principle of virtual work in Eq.~\ref{eq_forceinf}, yielding Eq.~\ref{eq_forcespecific}.

\begin{equation}
F_y = F_{l}\frac{dl}{dh}
\label{eq_forceinf}
\end{equation}
\begin{equation}
F_y  = \frac{\sqrt{3}F_{l} h^3}{2 \sqrt{4 b^2 h^4 - h^6}}
\label{eq_forcespecific}
\end{equation}

\noindent For the specific case of constant stiffness, that is $F_{l}  = k \Delta l$, with $k$ the so-called spring constant, and $\Delta l$ is the length difference between the distorted length $l$ and undistorted length $l_0$, the thrust force is given by Eq.~\ref{eq_forceconstk}.

\begin{equation}
F_y  = \frac{k h \left[ 2\sqrt{3} \left(c - l_0\right) h^2 + 3 \sqrt{\left(4 b^2 - h^2\right) h^4}\right]}{4\sqrt{\left(4 b^2 - h^2\right) h^4}}
\label{eq_forceconstk}
\end{equation}

\noindent To illustrate the thrust profile, {\textcolor{black}{the specific case of single pin knee joint, i.e., $p=q=0$ recovering the simplicity of the $S-1$ configuration, and equal leg length segments ($a=b$) in Eq.~\ref{eq_forceconstk} can be analysed.}} The spring in this robot will be an elastic material, which is assumed to only apply a load to the system when it extends beyond its natural length, and no load when it is compressed. This is modelled in the system using a Heaviside step function $\text{H} \left(\Delta l\right) $, collapses the thrust force to zero for cases where the spring is compressed.

\begin{figure} 
\centerline{\includegraphics[width=3.0in]{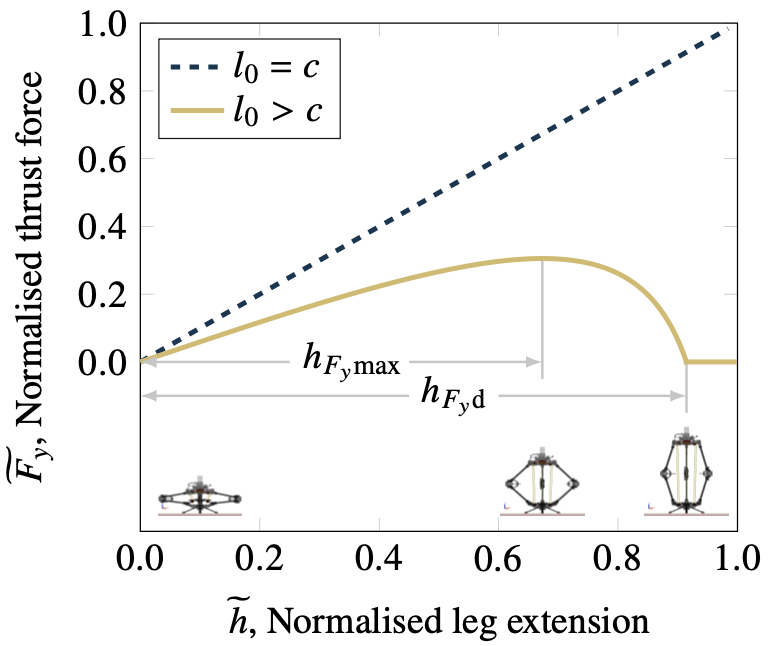}}
\caption{Thrust force vs. leg extension normalised to maximum values of $l_0 = c$, for equal leg length $a=1, k=1$. The curves illustrate two relation cases for  $l_0$ and $c$. For the sake of illustration $l_0 - c=0.7$ is used to cover the case $l_0 > c$. For $l_0 = c$, the force inversion observed in the hinge rhomb linkage is recovered. Conversely for $l_0 > c$ the behaviour of nonlinear spring is created.}
\label{fig_thrustprof}
\end{figure}

The resulting profile of $F_y$ is shown in Fig.~\ref{fig_thrustprof} for some combinations of $l_0$ and $c$, with $a$ and $k$ equal to one. From the figure it is observed that the thrust force peaks before full distension of the elastic element. In fact, for the condition of $c < l_0$ and $a > \sqrt{\frac{\left(c - l_0\right)^2}{3}}$, the thrust force peaks at (Eq.~\ref{eq_forcepeak})

\begin{equation}
h_{F_y\text{max}}  = 2 \sqrt{a^2 - \left[\frac{a^4 \left(c - l_0\right)^2}{3} \right]^{\frac{1}{3}}}
\label{eq_forcepeak}
\end{equation}

\noindent The linkage height at full spring distension, $h_{F_y\text{d}}$, is defined by Eq.~\ref{eq_hdistension} 

\begin{equation}
h_{F_y\text{d}} = 2 \sqrt{\frac{3 a^2 - \left(c - l_0\right)^2}{3}}
\label{eq_hdistension}
\end{equation}

\noindent Note in Fig.~\ref{fig_thrustprof} that for $l_0>c$ at $h_{F_y\text{d}}$ the force flow is null, which in practice implies minimum stress in joints. On the other hand, the closer the values of $l_0$ and $c$ are, the greater the force at $h_{F_y\text{d}}$, which result in backlash in joints when inertial effects are included. Backlash is a source of mechanical failure due to mechanical fatigue in joints as a result of the unreleased thrust energy circulating within the conversion mechanism. These two conditions call for trade-offs in the design to maintain acceptable performance as well as practical functionality as discussed in the following sections. 

\subsection{Dynamic modelling}
\label{subsec:dyn}

{\textcolor{black}{In this subsection the equations of motion for the system during the decompression phase up to the instant of take-off are formulated.}} Assume the geometric positions of gravity centres are given by Eqs.~\ref{eqn:positions}.

\begin{align}
\begin{split}
\bm{r}_{2,GC} &= \frac{a}{2}\cos \theta \: \uvec{i}+ \frac{a}{2}\sin \theta \:\uvec{j}\\
\bm{r}_{3,GC} &= a \cos \theta \:\uvec{i} +  \left(a \sin \theta + p \right) \:\uvec{j}\\
\bm{r}_{4,GC} &= \frac{a}{2}\cos \theta \:\uvec{i} +  \left(\frac{3}{2} a \sin \theta + 2p \right) \:\uvec{j}\\
\bm{r}_{5,GC} &= 0 \:\uvec{i} +  h \:\uvec{j}
\end{split}
\label{eqn:positions}
\end{align}

\noindent {\textcolor{black}{The velocities of the gravity centres in Eqs.~\ref{eqn:velocity} are derived from the geometric positions.}}

\begin{align}
\begin{split}
\dot{\bm{r}}_{2,GC} &= -\frac{a}{2}\dot{\theta} \left( \sin \theta \:\uvec{i} - \cos \theta \:\uvec{j}\right)\\
\dot{\bm{r}}_{3,GC} &= -a\dot{\theta} \left( \sin \theta \:\uvec{i} - \cos \theta \:\uvec{j}\right)\\
\dot{\bm{r}}_{4,GC} &= -\frac{a}{2}\dot{\theta} \left( \sin \theta \:\uvec{i} - 3 \cos \theta \:\uvec{j}\right)\\
\dot{\bm{r}}_{5,GC} &= 0 \:\uvec{i} + 2 a \cos \theta \dot{\theta} \:\uvec{j}
\end{split}
\label{eqn:velocity}
\end{align}

\noindent {\textcolor{black}{The Lagrangian mechanics approach is used to derive the governing equations of motion.}} The Lagrangian, $L \equiv T - V$, is derived from the potential of the conservative forces of the system. From Eqs.~\ref{eqn:velocity}, and from the substitution variables $\mathcal{M}_i$ defined in Eqs.~\ref{eqn:eqlabel} used to simplify equation notations, the total kinetic energy  $T$ can be expressed by Eq.~\ref{eq_kine}, with $I_1$ and $I_2$ as mass moments of inertia.

\begin{align}
\label{eqn:eqlabel}
\begin{split}
 \mathcal{M}_1 &= m_4+2 m_5
\\
\mathcal{M}_2 &= m_2 + 4 m_3 + 5 m_4 + 8 m_5
\\
\mathcal{M}_3 &= m_2 + 3 m_4 + 2 m_3 + 4 m_5
\\
\mathcal{M}_4 &= m_3 + 2 m_4 + 2 m_5
\end{split}
\end{align} 

\begin{equation}
T = \frac{a^2}{8}\left[4  \mathcal{M}_1 \cos 2\theta + \mathcal{M}_2 \right] \dot \theta^2 + \frac{1}{2} \left(I_1 + I_2 \right) \dot \theta^2
\label{eq_kine}
\end{equation}

\noindent From Eqs.~\ref{eqn:positions}, the total potential energy $V$ is given by Eq.~\ref{eq_kine}, with $g$ as the gravitational acceleration.

\begin{equation}
V = \frac{1}{2} a g \mathcal{M}_3 \sin \theta + p g \mathcal{M}_4
\label{eq_pote}
\end{equation}

\noindent As not all the forces acting on the system are derivable from a potential, the Lagrange's equation can be written as in Eq.~\ref{eq_genlagrangian}.

\begin{equation}
\frac{d}{d t}\frac{\partial L}{\partial \dot \theta} - \frac{\partial L}{\partial \theta} = Q
\label{eq_genlagrangian}
\end{equation}

\noindent In this equation $Q$ is the generalised forces term in Eq.~\ref{eq_genforce} given by the thrust force and non-conservative damping torques for frictional losses with characteristic Coulomb coefficient $\mu_C$.

\begin{equation}
Q = F_y \frac{\partial h}{\partial \theta} - \mu_C \text{sgn} \left(\dot{\theta}\right)
\label{eq_genforce}
\end{equation}

\noindent {\textcolor{black}{Finally, by solving Eq.~\ref{eq_genlagrangian} for $\ddot \theta$ and after further mathematical manipulation, Eq.~\ref{eq_lagrangiantotal} presents the Dynamical Model governing equation, hereinafter referred to as DM}}.

\begin{align}
\begin{split}
\ddot \theta = & \frac{4 \mathcal{M}_1 a^2 \sin 2 \theta}{a^2 \left(4 \mathcal{M}_1 \cos 2\theta+\mathcal{M}_2\right) + 4 \left( I_1+I_2\right)}\dot \theta^2 - \\
& \frac{2 a \cos \theta \left(g \mathcal{M}_3 - 4 F_y\right) + 4 \mu_C  \text{sgn} \left(\dot{\theta}\right)}{a^2 \left(4 \mathcal{M}_1 \cos 2\theta+\mathcal{M}_2\right) + 4 \left( I_1+I_2\right)}\\
\end{split}
\label{eq_lagrangiantotal}
\end{align}

\begin{figure*}[!ht]
\centerline{\includegraphics[width=3.0in]{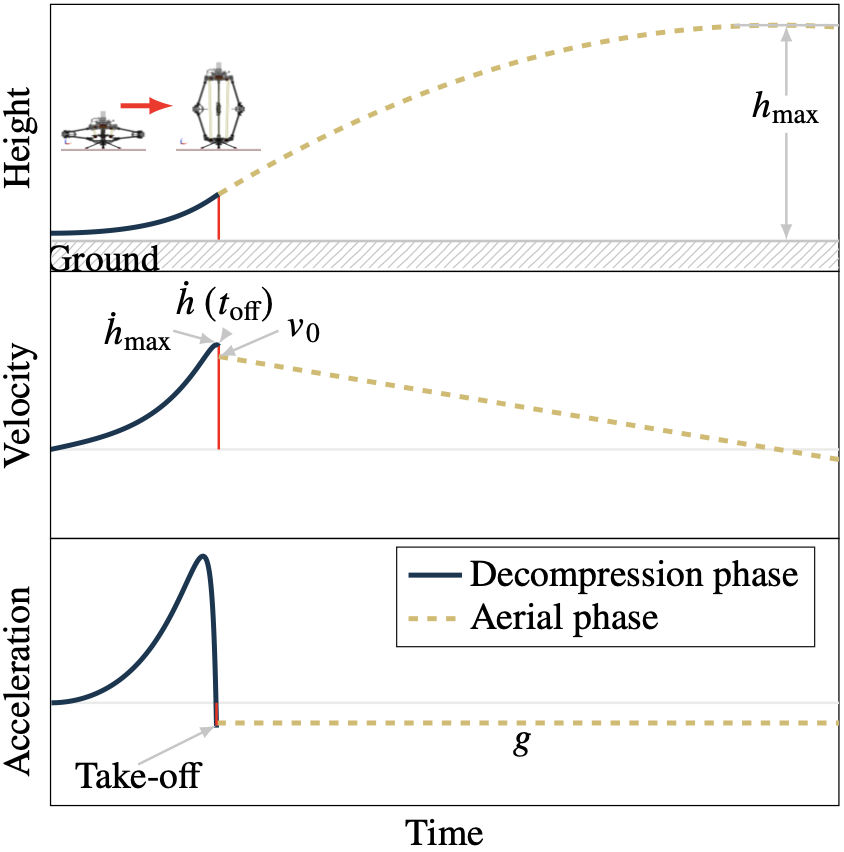}}
 \caption{Characteristic kinematic curves of a squat jump. Note that $v_0<\dot h\left(t_{\text{off}}\right)$; after reaching maximum velocity $\dot h_{\text{max}}$, the effect of gravity starts to slow down the motion but the upward movement continues. When the force due to body acceleration equals the weight of the foot, the whole linkage finally moves together and take-off occurs. The maximum flight height is denoted by $h_{\text{max}}$.}
  \label{fig_kcurves}
\end{figure*}

The dynamical model governing equation derived so far is meaningful only within the linkage decompression phase $0 \leq t \leq t_{\text{off}}$ ending at take-off. After this time, the aerial phase takes place. Also note that the governing equation is independent of $m_1$ as this remains static, in contact with the floor during the linkage decompression phase. Thus, the issue at hand is now to derive an equation for the take-off velocity that includes $m_1$ in order to simulate the jump trajectory, as shown in Fig.~\ref{fig_kcurves}. {\textcolor{black}{The first and second derivatives of Eq.~\ref{var_h} with respect to time yield the linear velocity and acceleration in Eq.~\ref{eq_hdot} and Eq.~\ref{eq_hddot} respectively.}}

\begin{align}
\dot h &= 2 a \cos \theta \dot \theta \label{eq_hdot} \\
\ddot h &= 2 a \cos \theta \ddot \theta - 2 a \sin \theta \dot \theta^2 \label{eq_hddot}
\end{align}

\noindent During the time the linkage is decompressing moving upwards, the mass in motion is $m_T - m_1$ with the total mass given by Eq.~\ref{eq_totmass}. Under this condition, the maximum value of Eq.~\ref{eq_hdot} is reached when the value of Eq.~ \ref{eq_hddot} is zero.

\begin{equation}
m_{T}=\sum_{i=1}^{5} m_i
\label{eq_totmass}
\end{equation}

\noindent After the total mass reaches its maximum velocity ($\dot h_{\text{max}}$), the effect of gravity starts to slow down the motion but the upward movement continues while the robot is still on the ground. The instant of take-off, $t_{\text{off}}$, occurs when the ground reaction force in Eq.~\ref{eq_toget} equals zero. 

\begin{equation}
F_N =  \left(m_T - m_1\right) \ddot h + \left(m_T - m_1\right)  g + m_1 g
\label{eq_toget}
\end{equation}

\noindent And at this instant the velocity of the centre of mass is defined as $\dot h_{\text{take-off}}$. Immediately after take-off the velocity of the centre of mass is obtained by assuming conservation of linear momentum:   

\begin{equation}
v_0 = \frac{m_{T} - m_1}{m_{T}} \dot h\left(t_{\text{off}}\right)
\label{eq_consmom}
\end{equation}

\noindent Note that for a massless foot ($m_1 = 0$), $v_0 =\dot h\left(t_{\text{off}}\right)$; in any other case some of the momentum is transferred to the foot.

\noindent Fig.~\ref{fig_kcurves} illustrates the jump dynamics. The decompression phase is modelled using Eqs.~\ref{eq_hdot} and Eq.~\ref{eq_hddot}, with the position data obtained from numerical integration of Eq.~\ref{eq_lagrangiantotal}. The aerial phase is treated as a ballistic trajectory in a gravity field with take-off velocity given by Eq.~\ref{eq_consmom}.

\section{Laboratory Demonstrator Design}
\label{sec:proto}

{\textcolor{black}{This investigation set out to develop a minimally actuated linkage platform for planetary jumping robots. In this section, the design of a laboratory demonstrator used to assist the evaluation of the proposed linkage approach is discussed. It is known from Section.~\ref{sec:mechdesign} that the proposed linkage has one translational DOF and is intended to develop linear momentum for a jump. It is worth mentioning that the take-off, aerial, and landing phases, each pose specific requirements for a practical jumping robot explorer. In practice, traverse mode would require thrust force vectoring, may require robot attitude control during aerial phase, and self-righting capability would be essential in preparation of a subsequent jump. Acknowledging this and that other robot design elements stem from the characteristics of the linkage dynamics, such as foot design and thrust force vectoring approach, the scope of the design of the laboratory demonstrator at hand is limited to the take-off development stage; specifically, the fundamental jumping ability metric, the potential-to-kinetic conversion efficiency is discussed in this section. Additionally, from Section.~\ref{sec:mechdesign}, note that the functional difference between the plates $m_2$ and $m_5$ in Fig.~\ref{fig_cad} is defined by the mass in contact with the floor at a specific time, i.e., the foot could be either plate according to its capability to generate thrust force. This characteristic of the linkage enables reversible jumping robot designs and simplistic robot self-righting approaches such as those presented by~\cite{Armour_2007,armour2010biologically}, which in practice are desirable for resilient locomotion. The reversible jumping robot approach will be preferred in subsequent iterations of the CLOVER robot. However, for the laboratory demonstrator design presented herein, the plates are predefined as foot and head in favour of a simple compression and latching mechanism for the linkage efficiency tests.}}

\noindent The overall design of the CLOVER robot {\textcolor{black}{laboratory demonstrator}} presented here embodies trade-offs between materials and components availability towards the intended desired functionality. Firstly, commercial off-the-shelf components and materials are preferred as a way to find rapid solutions to test fundamental design routes. In this tenor, the legs of the {\textcolor{black}{laboratory demonstrator}} are built with commercial carbon fibre box profile, and custom-made parts with PolyLactic Acid (PLA) through additive manufacturing. The selection of the carbon fibre box profile was made on the basis of bending resistance, light weight, and adequate subjection for leg attachments. For the CLOVER robot {\textcolor{black}{laboratory demonstrator}} TheraBand resistant band is identified as suitable energy storage drive because of its high snap resistance and availability material sizes and broad range of stiffness values (~\cite{patterson2001material}).

\noindent The experimental force-stretch curve for uniaxial elongation of a stripe of TheraBand Latex used in the CLOVER robot {\textcolor{black}{laboratory demonstrator}}, shows a characteristic hyperelastic behaviour of rubber-like materials. Rubber-like elasticity is suitably modelled with the Gaussian theory, which is a statistical-mechanics-based equation of state in Eq.~\ref{eq_gaussian}, where $\mathcal{C}_0$ is a material constant, $T$ is the temperature of the material, and deformation ratio $\lambda = \frac{l}{l_0}$. Although Gaussian theory breaks down at higher strains, it highlights the insightful relationship between force and temperature in rubber-like materials, i.e., the force is proportional to temperature. Therefore, temperature variations equally affect the restoring force. However, temperature and force are dependent variables of state in rubber-like materials. For example, a sudden deformation of the material that may render heat transfer from/to the environment negligible, i.e., quasi-adiabatic stretching, produces changes in the material temperature. In this way, if the material is suddenly stretched its temperature increases; conversely if it is suddenly distended its temperature decreases.

\begin{equation}
F_{l,G} = \mathcal{C}_0 T \left(\lambda-\frac{1}{\lambda^2}\right)
\label{eq_gaussian}
\end{equation}

\noindent TheraBand band stretching is a slow process in the CLOVER robot, which implies heat transfer from the band to the environment. Under this condition, an isothermal process taking place in the material is a reasonable assumption yielding force variation a sole function of the stretch. Band distension on the other hand, occurs in a fraction of a second enhancing the thermoelastic effect of the material. By comparing the thermoelasticity of a rubber band for a range of stretch values of interest for this research, i.e., $1\leq\lambda<3$, this supposes a temperature change of less than $1\si{\kelvin}$~(\cite{pellicer2001thermodynamics}). Furthermore, from Fig~\ref{fig_cad} is evident that linkage decompression involves rotational movement of the anchorage points of the band that in turn enhance heat transfer through forced convection. This mitigates the effect of thermoelastic force reduction during distension.

\noindent A better fit of the experimental values is achieved with the use of the phenomenological Mooney-Rivlin model for hyperelastic materials. The Mooney-Rivlin equation for the uniaxial deformation force-stretch is given by Eq.~\ref{eq_mooney}. The values of $\mathcal{C}_1$ and $\mathcal{C}_2$ are found through statistical curve fitting and reported in Table \ref{Table_theraband}. {\textcolor{black}{The Mooney-Rivlin model is used in the subsequent analysis.}}

\begin{equation}
\frac{F_{l,MR}}{A_0} = 2\mathcal{C}_1\left(\lambda-\frac{1}{\lambda^2}\right)+2\mathcal{C}_2\left(1-\frac{1}{\lambda^3}\right)
\label{eq_mooney}
\end{equation}

\noindent The energy conversion efficiency in Eq.~\ref{eq_efficiency}, is used to assess the {\textcolor{black}{laboratory demonstrator}} performance in transforming the potential energy into kinetic energy.

\begin{equation}
\eta = \frac{\mathcal{E}_K}{\mathcal{E}_P}\times 100 \%
\label{eq_efficiency}
\end{equation}

\noindent The potential energy\footnote{From Eq.~\ref{eq_mooney} and from the general definition of potential energy stored in a lossless flexile material  $\mathcal{E}_P = \int_{lo}^{lf} F_{l} dl$} stored in the hyperelastic material is given by Eq.~\ref{eq_potential}. The energy contained in the storage drive is ultimately released and transformed into various physical phenomena such as mechanical movement, heat, vibrations, etc. A key design objective in the design of jumping robots is the reduction of irreversibilities to transform most of the potential energy into kinetic energy available for the jump.

\begin{equation}
\mathcal{E}_P = \frac{A_0 l_0}{\lambda^2}\left(\lambda - 1 \right)^2 \left[ \mathcal{C}_1 \lambda \left(\lambda + 2 \right) + 2 \mathcal{C}_2 \lambda + \mathcal{C}_2 \right] \mid \lambda \geq 1
\label{eq_potential}
\end{equation}

\noindent Eq.~\ref{eq_kineticenergy} shows that the kinetic energy at take-off is given by its total mass $m_{T}$ and initial take-off velocity $v_0$.

\begin{equation}
\mathcal{E}_K = \frac{m_{T}}{2} v_0^2
\label{eq_kineticenergy}
\end{equation}

\noindent {\textcolor{black}{The following subsections \ref{subsect:compression} and \ref{subsect:docking} discuss the compression and docking mechanism approaches implemented in the CLOVER robot laboratory demonstrator}}.

\begin{table}[ht]
\centering
\caption{Fit model coefficients for the Mooney-Rivlin equation in Eq.~\ref{eq_mooney} representing the force-stretch curve of the strip of TheraBand\textsuperscript{\textregistered} Latex, $A_0=7 \si{\milli\metre\squared}$, at room temperature $T=296 \si{\kelvin}$. With $\mathcal{C}_0$ goodness of fit: root mean squared error $0.18$, R-Square $0.97$. With $\mathcal{C}_1$ and $\mathcal{C}_2$ goodness of fit: root mean squared error $0.03$, R-Square $0.99$.}
\label{Table_theraband}
\begin{tabular}[t]{lc}
\toprule
 Coefficient&Value\\
\midrule
$\mathcal{C}_0$&$47.94\times 10^{-4}$~$\si{\newton\per\kelvin}$\\
$\mathcal{C}_1$&$68.88~\si{\kilo\pascal}$\\
$\mathcal{C}_2$&$73.61~\si{\kilo\pascal}$\\
\bottomrule
\end{tabular}
\end{table}%

\subsection{Compression mechanism design approach}
\label{subsect:compression}  

\begin{figure*} 
\centerline{\includegraphics[width=4.0in]{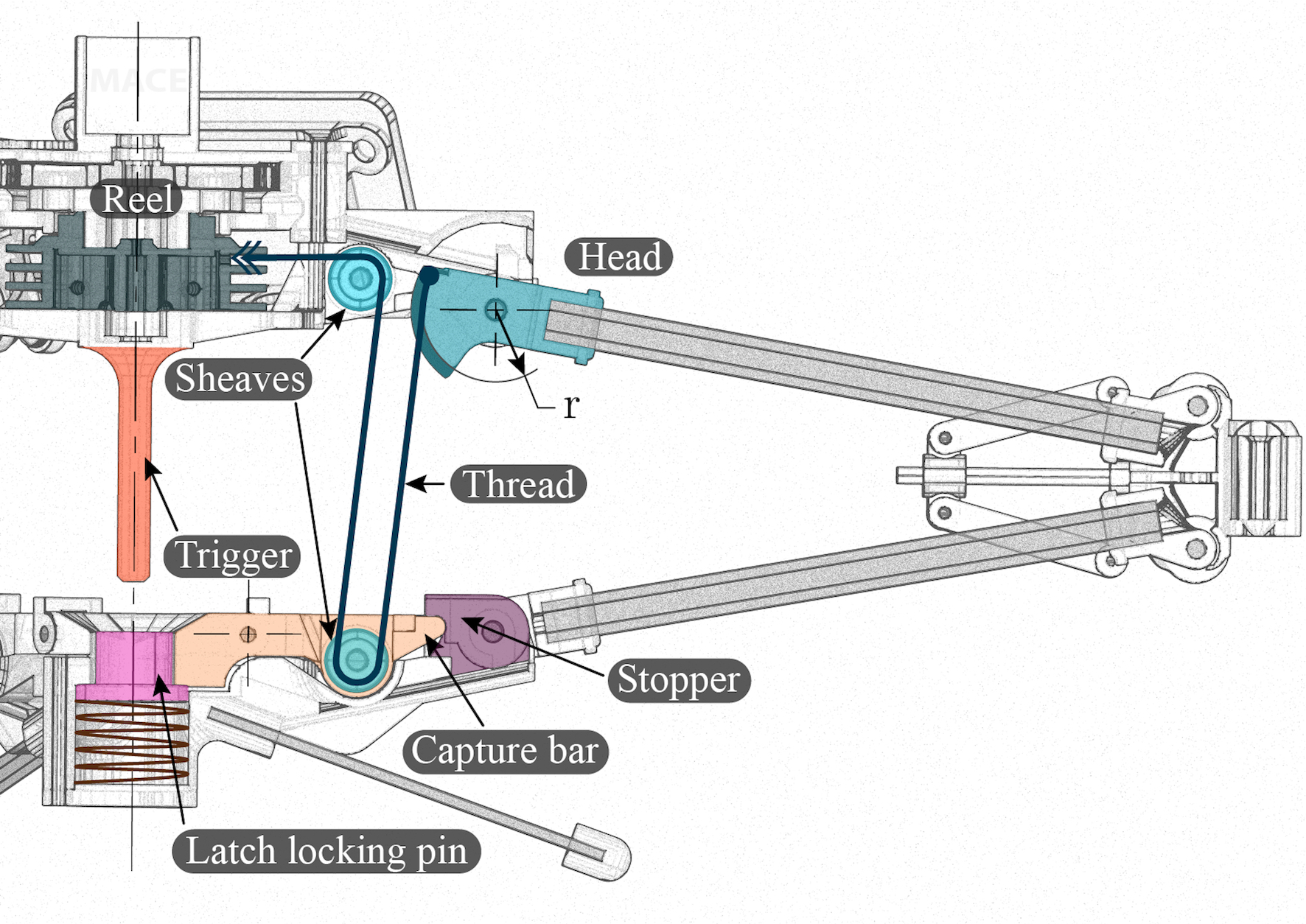}}
\caption{Compression mechanism and release latching mechanism approaches.}
\label{fig_pulley}
\end{figure*}

The compression mechanism shown in Fig.~\ref{fig_pulley}, consists of a compound pulley with mechanical advantage and velocity ratio of two. The extremes of the thread are anchored to the head of each leg and to the reel. A lightweight strong Kevlar thread passes through two sheaves, one located in the upper body and another in the lower body. Note that the head of the leg is a segment of a sheave of radius $r$ as shown in the figure. This shape offers two main advantages to the design: firstly, the initial tension, developed by the reel on the thread when this is wound, eases the compression process by ensuring outward flexion of the leg linkage acting like a class 1 lever (pin acting as a fulcrum), and secondly the resulting moment can ease the compression by enabling a least effort path. 

\begin{figure}
\centerline{\includegraphics[width=3.0in]{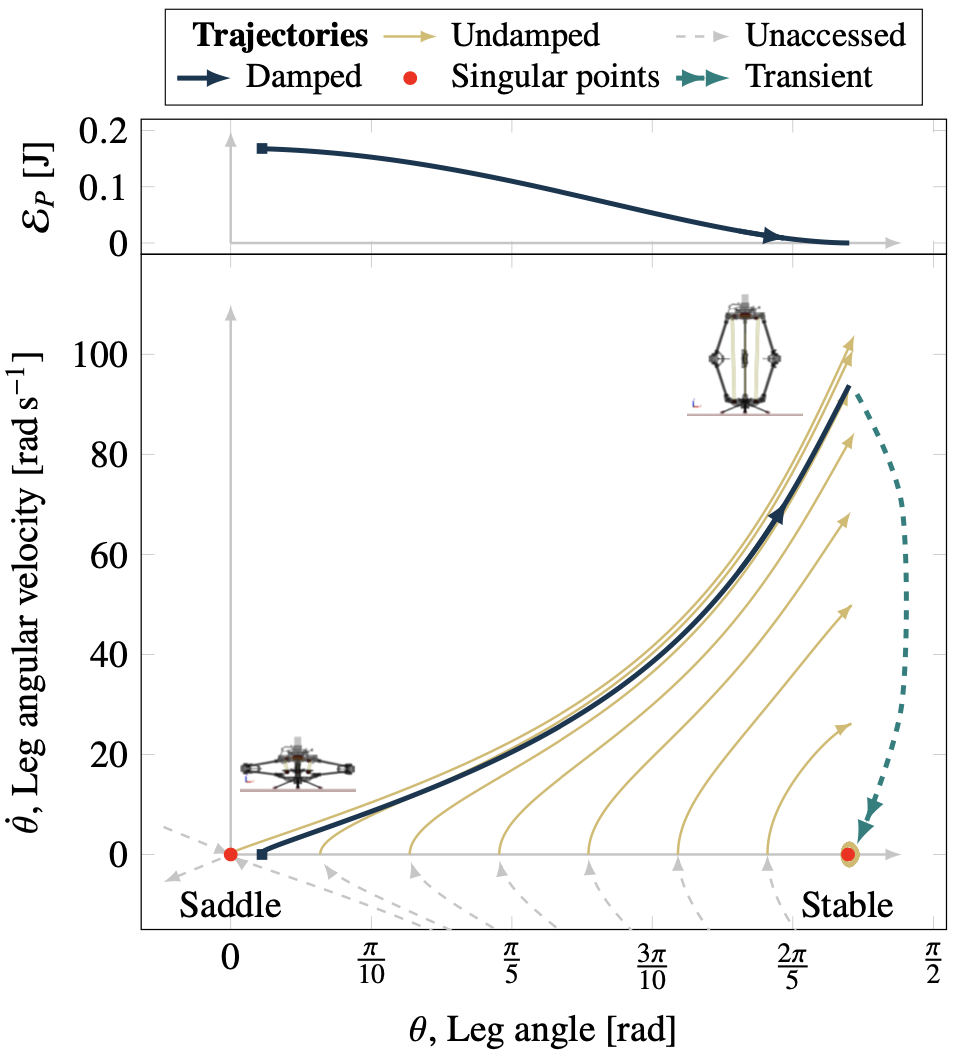}}
\caption{Phase portrait of the trajectories of the dynamical system for the parameters reported in Tables~\ref{Table_theraband} and \ref{Table_proto}. The damped trajectory characterising the {\textcolor{black}{laboratory demonstrator}} moves from a high potential energy state located near an unstable singular point, to a low potential energy state during linkage decompression. If the rotational kinetic energy ($\mathcal{E}_P$ converted to $\mathcal{E}_K$) in the linkage is damped out, then the trajectory of the dynamical model will move towards the stable singular point and remain idle afterwards. Undamped trajectories, i.e., $\mu_C = 0$, are shown for reference.}
\label{fig_port}
\end{figure}

\noindent The compression process gradually stores energy in the Theraband by the action of an electric motor. {\textcolor{black}{From the solutions of Eq.~\ref{eq_lagrangiantotal}, the phase portrait of the trajectories of the DM can be traced in Fig.~\ref{fig_port}.}} As the robot is compressed (reducing $\theta$) the state moves towards a singular point of the saddle point. Physically, the saddle point illustrates that the knee can open as intended (positive values of $\theta$) or can open inverted (negative values of $\theta$), the latter being considered as a failure mode. For this reason, practical designs should avoid being compressed to values of $\theta$ close to zero. 

\noindent When the stored elastic potential energy is released from the spring the system follows an equi-energetic trajectory; examples of release trajectories from various initial leg angles $\theta_0$ are shown in Fig.~\ref{fig_port}. 

Another interesting feature shown in the phase portrait is a singular point of the centre type in the vicinity of $\pi/2$; the location of this stable manifold in the phase portrait is related to the design of $F_y$ requiring $l_0 - c > 0$ for thrust force peak before full distension of the TheraBand band. In other words, the stable manifold corresponds to the rest leg angle at $\mathcal{E}_P=0$.

\noindent In a damped system, the trajectory of the dynamical model will reach equilibrium position by traversing equi-energetic trajectories. In a particular practical case, for angles near $\theta=0$, it is common that the thrust force is lower than static friction in joints impeding the decompression. For this reason, a practical design should avoid proximity to $\theta=0$. Acknowledging this is fundamental in the design of the docking mechanism discussed below.

\subsection{Docking mechanism design approach}
\label{subsect:docking}

Some examples of common energy release mechanism found in current jumping robots are overrunning clutches, snail cams, locking mechanisms, and detaching idler gears. In order to meet the sub-kilogram scale and low power consumption goals of the CLOVER robot design, the most suitable approach is to use a locking mechanism. The locking mechanism is a robust, simple, and small size solution for energy release. Fig.~\ref{fig_pulley} illustrates the fundamental components of the designed latching mechanism. The figure shows the docking configuration wherein the capture bar is firmly anchored by the stopper and latch locking pin. This configuration enables the operation of the compression mechanism. As the compression of the linkage evolves, the trigger gradually comes near to the latch locking pin, eventually pushing it releasing the capture bar; at that point, the decompression of the linkage starts. The length of the trigger establishes the initial value of $\theta$. From the previous discussion of the phase portrait of the trajectories of the DM, the length of the trigger must ensure that $\theta>0$ or if the system is highly damped, that the mechanism reaches sufficient thrust force as to overcome frictional forces in joints. The main advantages of the latching mechanism developed for the analysis presented herein is its simplicity and null friction losses after undocking. Nonetheless this solution enables untethered operation, it is impractical for autonomous application because it requires manual re-docking. For the analysis at hand, i.e., linkage system energy conversion efficiency, this approach is valid.

\section{Assessment of Laboratory Demonstrator}

The first task in this section is to analyse the theoretical characteristic evolution of the distension force $F_l$ and thrust force $F_y$. For this purpose, the CLOVER robot design parameters are presented in Table~\ref{Table_proto} for the evaluation of the dynamical model. An important characteristic of the design is that $F_y$ starts at a value different from zero. The {\textcolor{black}{laboratory demonstrator}} has been designed with this characteristic to prompt and ease decompression since the squat position is located near an equilibrium point as explained by the dynamical model governing equation response, Eq.~\ref{eq_lagrangiantotal}. The phase portrait of the trajectories of the dynamical model within the region of interest, $0 \leq \theta \leq \pi/2$, exhibit a saddle point at the origin of the axis and a stable point approximately at 1.3 \si{\radian} corresponding to the leg angle $\theta$ in the standing position.

\begin{table}[!ht]
\centering
\caption{{\textcolor{black}{Laboratory demonstrator}} parameters for each element in Fig.~\ref{fig_cad}. The total mass of the CLOVER robot is $75.3\times10^{-3}$ \si{\kilogram}.} 
\label{Table_proto}
\begin{threeparttable}
\begin{tabular}[t]{lcc}
\toprule
 &Value& Units\\
\midrule
$a$&$6.82\times10^{-2}$&\si{\metre}\\
$c$&$5.50\times10^{-2}$&\si{\metre}\\
$g$&$9.81$&\si{\metre\per\second\squared}\\
$I_1$&$6.28\times10^{-7}$&\si{\kilogram\metre\squared}\\
$I_2$&$6.28\times10^{-7}$&\si{\kilogram\metre\squared}\\
$l_0$&$8.50\times10^{-2}$&\si{\metre}\\
$m_1$\tnote{*}&$2.70\times10^{-3}$&\si{\kilogram}\\
$m_2$&$1.60\times10^{-3}$&\si{\kilogram}\\
$m_3$&$3.10\times10^{-3}$&\si{\kilogram}\\
$m_4$&$1.60\times10^{-3}$&\si{\kilogram}\\
$m_5\tnote{*}$&$16.10\times10^{-3}$&\si{\kilogram}\\
$p$&$0.70\times10^{-2}$&\si{\metre}\\
$q$&$0.50\times10^{-2}$&\si{\metre}\\
\bottomrule
\end{tabular}
    \begin{tablenotes}[flushleft]\footnotesize
      \item[*] One third of the actual part.
    \end{tablenotes}
\end{threeparttable}
\end{table}

\begin{figure} 
\centerline{\includegraphics[width=3.0in]{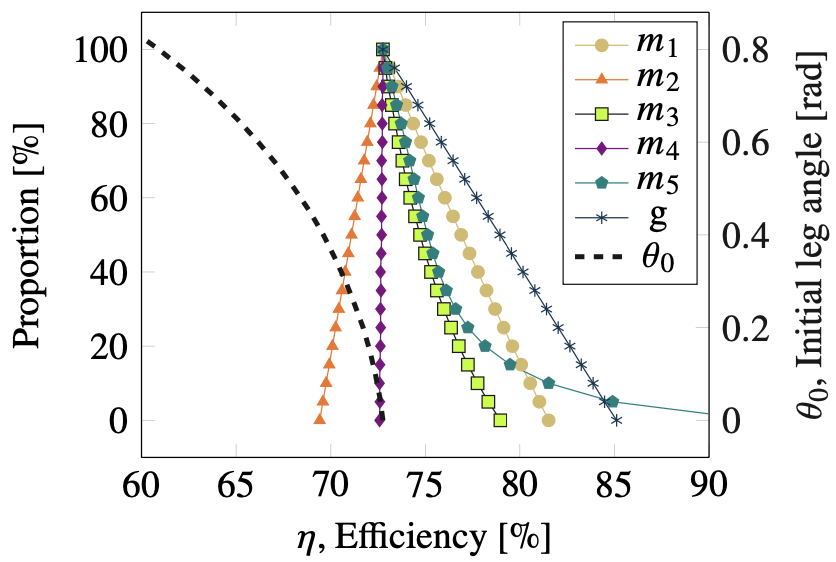}}
\caption{Undamped energy conversion efficiency sensitivity to {\textcolor{black}{laboratory demonstrator}} parameters. Proportions with respect to the values in Table~\ref{Table_proto}.}
\label{fig_sens}
\end{figure}

\noindent  All parameters and known initial conditions of the {\textcolor{black}{laboratory demonstrator}} design allow us to make estimations of {\textcolor{black}{its theoretical conversion efficiency in Eq.~\ref{eq_efficiency}, using the theory developed in Subsection~\ref{subsec:dyn}}}. The theoretical undamped energy conversion efficiency sensitivity to the {\textcolor{black}{laboratory demonstrator with}} parameters in Table~\ref{Table_proto}, is shown in Fig.~\ref{fig_sens}. The figure shows the  {\textcolor{black}{theoretical sensitivity of the demonstrator to single parameter variations}}. With the exception of $\theta_0$, each parameter is changed proportionally from $0$ to $100\%$ of their original value. {\textcolor{black}{In Fig.~\ref{fig_sens} is observed that the nominal characteristic undamped efficiency of the laboratory demonstrator, i.e., proportions at $100\%$ and $\theta_0 \rightarrow 0$, is $73\%$, which represents the maximum attainable value for the mass and geometric proportions of the demonstrator.}}

In the first instance, efficiency is mostly influenced by the value of gravity acting against the linear vertical acceleration, $\ddot h$; a reduction in $g$ value prompts an increase in $\eta$.  From the body masses, the foot mass $m_1$ and the upper mass $m_5$ are the most influential; a value increase reduces $\eta$.
Less detrimental variation on $\eta$ is observed for knee mass ($m_3$) increment, which supports the CLOVER knee design approach.

\noindent The initial leg angle, $\theta_0$, also affects the conversion efficiency as shown in Fig.~\ref{fig_sens}. Theoretical efficiency peaks at $\theta_0=0$ as this creates maximum Theraband extension which in turn represents the highest potential energy for a given geometry and mass configuration. For the sake of practical implementation, the value of $\theta_0=0$ in the CLOVER {\textcolor{black}{laboratory demonstrator}} is $0.066~\si{\radian}$, equivalent to $3.8^{\circ}$. This empirical value prevented linkage blockage near the saddle point due to lax tolerances in the manufacturing process. With this initial leg angle and the parameters reported in Table~\ref{Table_proto}, the value of $\mathcal{E}_P$ is estimated in $0.17 \si{\joule}$, and new maximum energy conversion efficiency of $72.5\%$.

\begin{figure}
    \centering
    \begin{minipage}{0.45\textwidth}
        \centering
        \includegraphics[width=0.9\textwidth]{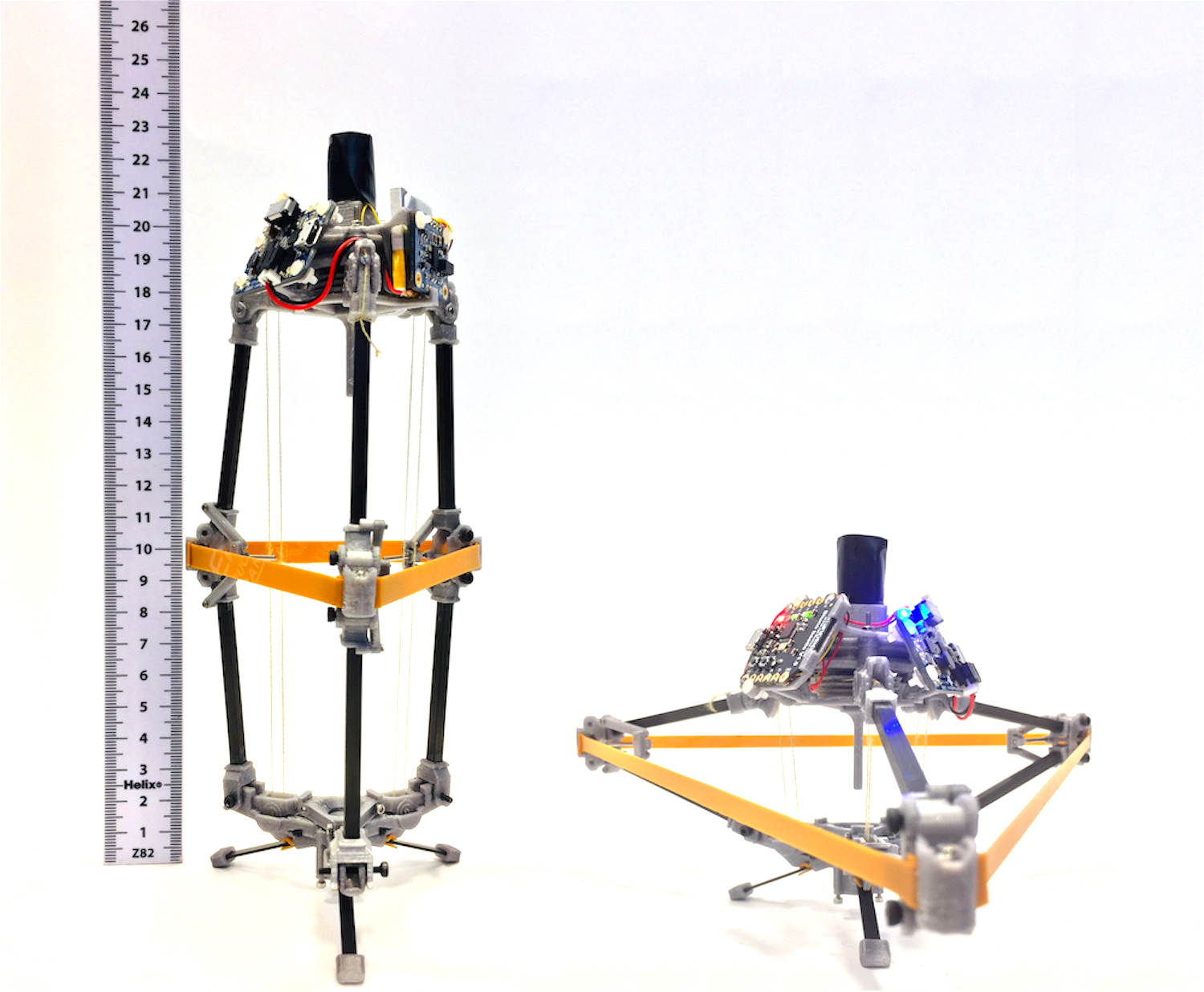} 
    \vspace*{-30ex} 
\begin{flushleft}
{\bf ~~~~A}
\end{flushleft}
\vspace*{24ex} 

    \end{minipage}\hfill
    \begin{minipage}{0.52\textwidth}
        \centering
        \includegraphics[width=1\textwidth]{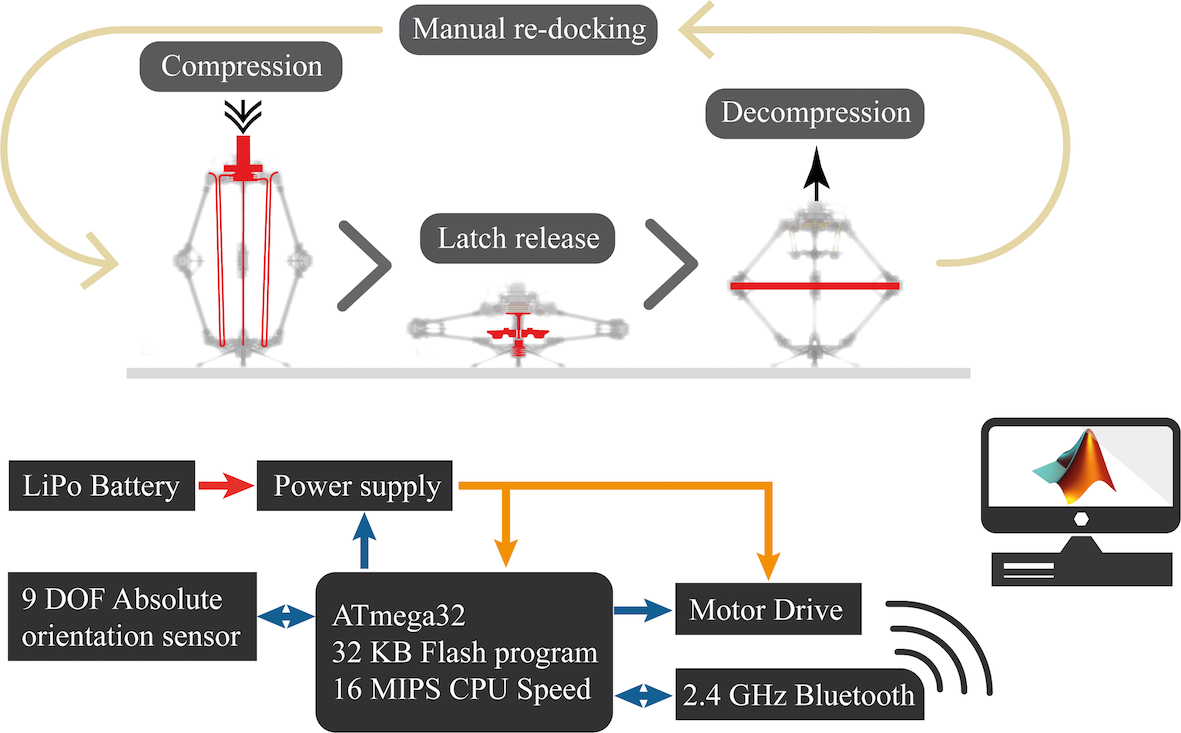} 
      \vspace*{-30ex} 
\begin{flushleft}
{\bf ~~~B}
\end{flushleft}
\vspace*{24ex} 

    \end{minipage}
    \caption{{\bf A}, Physical {\textcolor{black}{laboratory demonstrator}} in stand (idle) and squat position, operation sequence, and {\bf B} robot network architecture.}
    \label{fig_fdesign}
\end{figure}

\subsection{{\textcolor{black}{Laboratory Demonstrator Experimental}} Jumping Performance}

{\textcolor{black}{
Section~\ref{sec:proto} introduced the concept of energy conversion efficiency to evaluate jumping ability. In this section the fundamental jumping ability of the laboratory demonstrator is evaluated through the analysis of vertical jump tests. To this aim, the CLOVER robot demonstrator, a Bluetooth dongle, a computer with MATLAB software, and a video camera were used in the experiments. 
}}

{\textcolor{black}{
In order to reduce energy damping losses between the ground and the robot to approximate the theoretical behaviour discussed above, jump tests are carried out on a flat solid horizontal surface. The robot foot sat flat on the horizontal surface before each jump experiment. Serial port communication was established between the MATLAB and dongle to read streaming data from the robot. The camera was set up to capture a perspective view of the robot jump.
}}

{\textcolor{black}{
The research protocol started with the wireless connection between the robot and the dongle. The locking mechanism was manually assembled with the robot in idle stand position as shown in Fig.~\ref{fig_fdesign}~{\bf A}. Commands were sent through MATLAB to the robot to test successful communication. With the connection stablished and test commands executed by the robot, a wind command actuated the reel and compression of the linkage started for an experiment. Latch release happened eventually by the self-compression of the linkage ultimately yielding a jump. With the robot fully decompressed after a jump, manual re-docking prepared the robot for a subsequent jump. An illustration summarising this process is presented in Fig.~\ref{fig_fdesign}~{\bf B}.
}}

\begin{figure*}[!ht]
\centerline{\includegraphics[width=5.0in]{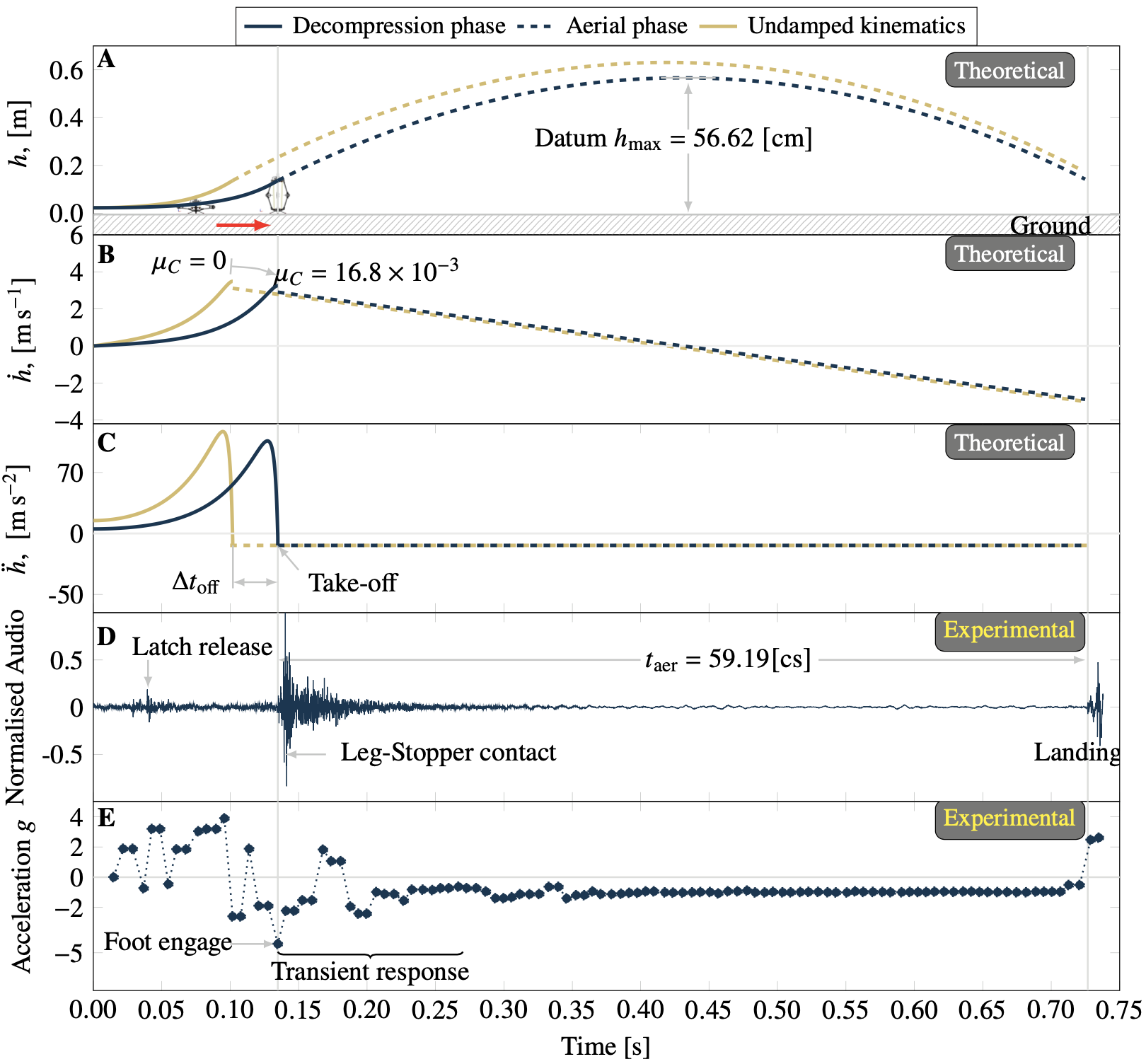}}
 \caption{{\bf A}--{\bf C} kinematic curves for the CLOVER robot {\textcolor{black}{laboratory demonstrator}}. After take-off, a transient response in the linkage is observed in {\bf E}. Ground contact (landing phase) is observed as a sudden signal transition in {\bf D} and {\bf E}. The onboard acceleration readings are bound by the accelerometer capability to approximately $\pm 4 \text{g}$.}
  \label{fig_kcurves2}
\end{figure*}

{\textcolor{black}{
The experimental evolution of the dynamics of the laboratory demonstrator was registered externally by the video camera, and onboard by the 9DOF absolute orientation sensor. The camera captured 240 video frames per second and 44 100 audio samples per second, while the orientation sensor captured 150 samples per second during the jump tests. The time-of-flight characteristic to the vertical jump is a conspicuous phenomenon (Fig.~\ref{fig_kcurves2}~{\bf A}-{\bf C}), which was effectively captured by media as well as the onboard orientation sensor as shown in Fig.~\ref{fig_kcurves2}~{\bf D} and Fig.~\ref{fig_kcurves2}~{\bf E} respectively. Notwithstanding, higher sampling of the acoustic data provided the best resource for the analysis. This data captured acoustic signatures related to well-known sequential stages in the experiment. In the recorded sequence shown in Fig.~\ref{fig_kcurves2}~{\bf D}, the first sound spike as seen from left to right, was caused by the latch release marking the start of the jump development. The next major sound transition in the sequence captured the foot engage marking the start of the aerial phase. Subsequently, a high amplitude sound spike produced by the leg-stopper contact (diagram in Fig.~\ref{fig_pulley}), was followed by a decreasing muffled sound evincing the transient dynamical response of the CLOVER robot. The final recorded sound spike demarcated the end of the aerial phase at robot-floor contact. Therefore, the characteristic time of the take-off phase of the vertical jump was 95~\si{\milli\second}, and the characteristic time lapse between foot engagement and landing floor contact, i.e., the time-of-flight, was 592~\si{\milli\second} with standard deviation of 59.7~\si{\milli\second} from ten observations.}}

\noindent {\textcolor{black}{The take-off velocity was inferred\footnote{From $v_0 = \frac{g t_{\text{aer}}}{2}$} from the time-of-flight $t_{\text{aer}}$ as 2.9 \si{\metre\per\second}, yielding an estimated maximum jump height\footnote{From $v_0 =\sqrt{2 g h_{\text{max}}}$} of 0.566 \si{\metre} verified by observations. The inferred take-off velocity was then used to estimate the characteristic mechanical system damping of the laboratory demonstrator, embodied by the Coulomb coefficient $\mu_C$ in the DM (Subsection~\ref{subsec:dyn}). The value of $\mu_C = 16.811\times 10^{-3}$ yielding the experimentally inferred value of $v_0$ was identified through its numerical iteration in the DM. The definition of $\mu_C$ different from zero defined in turn the new kinematic curves for the resulting damped DM shown in Fig.~\ref{fig_kcurves2}~{\bf A}-{\bf C}. This also set a take-off time difference $\Delta t_{\text{off}}$ between the damped and undamped DMs curves of 33 \si{\milli\second}, Fig.~\ref{fig_kcurves2}~{\bf C}. Interestingly, by matching the experimental take-off times to the damped DM as reported in Fig.~\ref{fig_kcurves2}~{\bf D} and~{\bf E}, the experimental decompression time (from latch release to take-off) is clearly shorter than in the DM curves (from time zero to take-off). This observed discrepancy of 40 \si{\milli\second} is originated by the operation of the docking mechanism (Fig.~\ref{fig_pulley}). When the trigger pushed the latch locking pin at the end of the linkage compression, a small amount of energy was stored in the helical compression spring that kept it in place. Just after latch release, i.e., when the capture bar-stopper geometric constrain is finally removed by the trigger action, part of the potential energy stored in the helical spring was transformed directly into kinetic energy of $m_5$ by the upward pressure exerted on the trigger by the latch locking pin. Finally, the estimated energy conversion efficiency of the {\textcolor{black}{laboratory demonstrator}} was computed with the potential energy $\mathcal{E}_P$ value from Eq.~\ref{eq_potential} with coefficients from Table ~\ref{Table_theraband}, and the inferred kinetic energy at take-off from Eq.~\ref{eq_kineticenergy} with the experimental take-off velocity value; the estimated energy conversion efficiency of the laboratory demonstrator was $63.1\%$, that is 9.4 percentage point decrease with respect to the maximum undamped theoretical case. Reference experiments carried out by \cite{hale2000minimally} on a 1.5\si{\kilogram} two-leg spring driven gear synchronised jumping robot realised a $70\%$ energy conversion efficiency. From friction in joints, parasitic vibrations, structural damping, to heat generation mostly from component collisions, various energy dissipation mechanisms contribute in detriment of $\eta$. However, joint friction losses quantified by $\mu_C$ in the leg design of the CLOVER robot demonstrator clearly dominate.}}

\section{Conclusions and Future Work}

This paper described the theoretical development and practical demonstration of a jumping robot. The study drew inspiration from the JPL planetary exploration robot, which utilised a nonlinear spring in order to increase elastic-to-kinetic energy transfer in comparison to a linear spring.

\noindent The present study extends this approach using a Sarrus-style linkage to constrain the system to a single translation degree of freedom without the use of gears, while still retaining the favourable nonlinear force profile. This offers a route for feasible integration into future planetary robots, where the risk of dust ingress into gears is prohibitively high. Linkage flexion direction yielding symmetric (mirrored) tension on the elastic material in the practical demonstration was assisted by a pulley-lever system, which provided compression capability while enabling redundant least effort paths of compression. The configuration of pulley-lever system is designed to convey the dynamical system from the vicinity of a singular point of the saddle type when in linkage compression, to a stable one in linkage extension through a mechanically suitable path. These mechanical elements were realised as a functional hopping robot {\textcolor{black}{laboratory demonstrator}}. The robot demonstrates 63\% potential-to-kinetic energy conversion efficiency, with a theoretical maximum of 73\% for this particular spring and linkage system with no dissipative losses.

\noindent The theoretical maximum is limited by the inertial component of the mechanism-ground interface, i.e., robot foot mass, and by the characteristic spring force-strain curve. Transformation of potential energy from the spring into kinetic energy of the linkage, is inherently related to the state of the mass configuration of the system at any specific time; this observation implies the existence of an optimum spring force-strain curve for maximum mechanical conversion efficiency for the masses and geometry of the linkage. These observations will be subject of future research.

\section*{Author Contributions}

All authors contributed to conception and design of the study. Robot {\textcolor{black}{laboratory demonstrator}} development, experimental setup, and data analysis carried out by AM. All authors contributed to manuscript revision, read, and approved the submitted version.

\section*{Funding}

The authors gratefully acknowledge the support of the Future AI and Robotics Hub for Space (FAIR-SPACE), EPSRC Grant code EP/R026092/1, and the University of Manchester Research Institute for funding this research. Additionally, we acknowledge support by the University of Manchester for providing Open Access funding.

\section*{Acknowledgments}
The authors would like to thank the University of Manchester for supporting this research.

\bibliographystyle{frontiersinSCNS_ENG_HUMS} 
\bibliography{clover}

\begin{thebibliography}{31}
\providecommand{\natexlab}[1]{#1}
\expandafter\ifx\csname urlstyle\endcsname\relax
  \providecommand{\doi}[1]{doi:\discretionary{}{}{}#1}\else
  \providecommand{\doi}{doi:\discretionary{}{}{}\begingroup
  \urlstyle{rm}\Url}\fi
\providecommand{\selectlanguage}[1]{\relax}
\providecommand{\bibAnnoteFile}[1]{%
  \IfFileExists{#1}{\begin{quotation}\noindent\textsc{Key:} #1\\
  \textsc{Annotation:}\ \input{#1}\end{quotation}}{}}
\providecommand{\bibAnnote}[2]{%
  \begin{quotation}\noindent\textsc{Key:} #1\\
  \textsc{Annotation:}\ #2\end{quotation}}

\bibitem[{Armour et~al.(2007)Armour, Paskins, Bowyer, Vincent, and
  Megill}]{Armour_2007}
Armour, R., Paskins, K., Bowyer, A., Vincent, J., and Megill, W. (2007).
\newblock Jumping robots: a biomimetic solution to locomotion across rough
  terrain.
\newblock \emph{Bioinspiration {\&} Biomimetics} 2, S65--S82.
\newblock \doi{10.1088/1748-3182/2/3/s01}
\bibAnnoteFile{Armour_2007}

\bibitem[{Armour(2010)}]{armour2010biologically}
Armour, R.~H. (2010).
\newblock \emph{A biologically inspired jumping and rolling robot}.
\newblock Ph.D. thesis, University of Bath
\bibAnnoteFile{armour2010biologically}

\bibitem[{Bai et~al.(2019)Bai, Zheng, Chen, Sun, and Hou}]{bai2019design}
Bai, L., Zheng, F., Chen, X., Sun, Y., and Hou, J. (2019).
\newblock Design and experimental evaluation of a single-actuator continuous
  hopping robot using the geared symmetric multi-bar mechanism.
\newblock \emph{Applied Sciences} 9, 13
\bibAnnoteFile{bai2019design}

\bibitem[{Batts et~al.(2016)Batts, Kim, and Yamane}]{batts2016untethered}
Batts, Z., Kim, J., and Yamane, K. (2016).
\newblock Untethered one-legged hopping in 3d using linear elastic actuator in
  parallel (leap).
\newblock In \emph{International Symposium on Experimental Robotics}
  (Springer), 103--112
\bibAnnoteFile{batts2016untethered}

\bibitem[{Bayliss and Langley(2003)}]{bayliss2003nuclear}
Bayliss, C. and Langley, K. (2003).
\newblock \emph{Nuclear decommissioning, waste management, and environmental
  site remediation} (Elsevier)
\bibAnnoteFile{bayliss2003nuclear}

\bibitem[{Budynas et~al.(2005)Budynas, Nisbett, and
  Tangchaichit}]{budynas2005shigley}
Budynas, R.~G., Nisbett, J.~K., and Tangchaichit, K. (2005).
\newblock \emph{Shigley's mechanical engineering design} (McGraw Hill New York)
\bibAnnoteFile{budynas2005shigley}

\bibitem[{Carpi et~al.(2011)Carpi, De~Rossi, Kornbluh, Pelrine, and
  Sommer-Larsen}]{carpi2011dielectric}
Carpi, F., De~Rossi, D., Kornbluh, R., Pelrine, R.~E., and Sommer-Larsen, P.
  (2011).
\newblock \emph{Dielectric elastomers as electromechanical transducers:
  Fundamentals, materials, devices, models and applications of an emerging
  electroactive polymer technology} (Elsevier)
\bibAnnoteFile{carpi2011dielectric}

\bibitem[{Chignoli et~al.(2021)Chignoli, Kim, Stanger-Jones, and
  Kim}]{chignoli2021humanoid}
Chignoli, M., Kim, D., Stanger-Jones, E., and Kim, S. (2021).
\newblock The mit humanoid robot: Design, motion planning, and control for
  acrobatic behaviors.
\newblock \emph{arXiv preprint arXiv:2104.09025}
\bibAnnoteFile{chignoli2021humanoid}

\bibitem[{Ellery(2005)}]{ellery2005environment}
Ellery, A. (2005).
\newblock Environment robot interaction the basis for mobility in planetary
  micro-rovers.
\newblock \emph{Robotics and Autonomous Systems} 51, 29--39
\bibAnnoteFile{ellery2005environment}

\bibitem[{Ellery(2015)}]{ellery2015planetary}
Ellery, A. (2015).
\newblock \emph{Planetary rovers: robotic exploration of the solar system}
  (Springer)
\bibAnnoteFile{ellery2015planetary}

\bibitem[{ESA(2021)}]{esalavatube}
[Dataset] ESA (2021).
\newblock Esa plans mission to explore lunar caves
\bibAnnoteFile{esalavatube}

\bibitem[{Hale et~al.(2000)Hale, Schara, Burdick, and
  Fiorini}]{hale2000minimally}
Hale, E., Schara, N., Burdick, J., and Fiorini, P. (2000).
\newblock A minimally actuated hopping rover for exploration of celestial
  bodies.
\newblock In \emph{Proceedings 2000 ICRA. Millennium Conference. IEEE
  International Conference on Robotics and Automation. Symposia Proceedings
  (Cat. No. 00CH37065)} (IEEE), vol.~1, 420--427
\bibAnnoteFile{hale2000minimally}

\bibitem[{Harvey(2007)}]{harvey2007russian}
Harvey, B. (2007).
\newblock \emph{Russian planetary exploration: history, development, legacy and
  prospects} (Springer Science \& Business Media)
\bibAnnoteFile{harvey2007russian}

\bibitem[{Ho et~al.(2017)Ho, Baturkin, Grimm, Grundmann, Hobbie, Ksenik
  et~al.}]{ho2017mascot}
Ho, T.-M., Baturkin, V., Grimm, C., Grundmann, J.~T., Hobbie, C., Ksenik, E.,
  et~al. (2017).
\newblock Mascot---the mobile asteroid surface scout onboard the hayabusa2
  mission.
\newblock \emph{Space Science Reviews} 208, 339--374
\bibAnnoteFile{ho2017mascot}

\bibitem[{Huang et~al.(2013)Huang, Li, and Ding}]{huang2013mobility}
Huang, Z., Li, Q., and Ding, H. (2013).
\newblock Mobility analysis part-1.
\newblock In \emph{Theory of Parallel Mechanisms} (Springer). 47--70
\bibAnnoteFile{huang2013mobility}

\bibitem[{Jung et~al.(2016)Jung, Casarez, Jung, Fearing, and
  Cho}]{jung2016integrated}
Jung, G.-P., Casarez, C.~S., Jung, S.-P., Fearing, R.~S., and Cho, K.-J.
  (2016).
\newblock An integrated jumping-crawling robot using height-adjustable jumping
  module.
\newblock In \emph{2016 IEEE International Conference on Robotics and
  Automation (ICRA)} (IEEE), 4680--4685
\bibAnnoteFile{jung2016integrated}

\bibitem[{Liljeback et~al.(2010)Liljeback, Pettersen, Stavdahl, and
  Gravdahl}]{liljeback2010controllability}
Liljeback, P., Pettersen, K.~Y., Stavdahl, {\O}., and Gravdahl, J.~T. (2010).
\newblock Controllability and stability analysis of planar snake robot
  locomotion.
\newblock \emph{IEEE Transactions on Automatic Control} 56, 1365--1380
\bibAnnoteFile{liljeback2010controllability}

\bibitem[{Maimone et~al.(2006)Maimone, Biesiadecki, Tunstel, Cheng, and
  Leger}]{maimone2006surface}
Maimone, M., Biesiadecki, J., Tunstel, E., Cheng, Y., and Leger, C. (2006).
\newblock Surface navigation and mobility intelligence on the mars exploration
  rovers.
\newblock \emph{Intelligence for space robotics} , 45--69
\bibAnnoteFile{maimone2006surface}

\bibitem[{Parslew et~al.(2018)Parslew, Sivalingam, and
  Crowther}]{parslew2018dynamics}
Parslew, B., Sivalingam, G., and Crowther, W. (2018).
\newblock A dynamics and stability framework for avian jumping take-off.
\newblock \emph{Royal Society open science} 5, 181544
\bibAnnoteFile{parslew2018dynamics}

\bibitem[{Patterson et~al.(2001)Patterson, Stegink~Jansen, Hogan, and
  Nassif}]{patterson2001material}
Patterson, R.~M., Stegink~Jansen, C.~W., Hogan, H.~A., and Nassif, M.~D.
  (2001).
\newblock Material properties of thera-band tubing.
\newblock \emph{Physical therapy} 81, 1437--1445
\bibAnnoteFile{patterson2001material}

\bibitem[{Pellicer et~al.(2001)Pellicer, Manzanares, Z{\'u}{\~n}iga, Utrillas,
  and Fern{\'a}ndez}]{pellicer2001thermodynamics}
Pellicer, J., Manzanares, J.~A., Z{\'u}{\~n}iga, J., Utrillas, P., and
  Fern{\'a}ndez, J. (2001).
\newblock Thermodynamics of rubber elasticity.
\newblock \emph{Journal of Chemical Education} 78, 263
\bibAnnoteFile{pellicer2001thermodynamics}

\bibitem[{Plecnik et~al.(2017)Plecnik, Haldane, Yim, and
  Fearing}]{plecnik2017design}
Plecnik, M.~M., Haldane, D.~W., Yim, J.~K., and Fearing, R.~S. (2017).
\newblock Design exploration and kinematic tuning of a power modulating jumping
  monopod.
\newblock \emph{Journal of Mechanisms and Robotics} 9, 011009
\bibAnnoteFile{plecnik2017design}

\bibitem[{Seeni et~al.(2010)Seeni, Sch{\"a}fer, and Hirzinger}]{seeni2010robot}
Seeni, A., Sch{\"a}fer, B., and Hirzinger, G. (2010).
\newblock Robot mobility systems for planetary surface
  exploration--state-of-the-art and future outlook: a literature survey.
\newblock \emph{Aerospace Technologies Advancements} 492, 189--208
\bibAnnoteFile{seeni2010robot}

\bibitem[{Smith et~al.(2020)Smith, Craig, Herrmann, Mahoney, Krezel, McIntyre
  et~al.}]{smith2020artemis}
Smith, M., Craig, D., Herrmann, N., Mahoney, E., Krezel, J., McIntyre, N.,
  et~al. (2020).
\newblock The artemis program: An overview of nasa's activities to return
  humans to the moon.
\newblock In \emph{2020 IEEE Aerospace Conference} (IEEE), 1--10
\bibAnnoteFile{smith2020artemis}

\bibitem[{Truong et~al.(2019)Truong, Phan, and Park}]{truong2019design}
Truong, N.~T., Phan, H.~V., and Park, H.~C. (2019).
\newblock Design and demonstration of a bio-inspired flapping-wing-assisted
  jumping robot.
\newblock \emph{Bioinspiration \& biomimetics} 14, 036010
\bibAnnoteFile{truong2019design}

\bibitem[{Ulamec et~al.(2011)Ulamec, Kucherenko, Biele, Bogatchev, Makurin, and
  Matrossov}]{ulamec2011hopper}
Ulamec, S., Kucherenko, V., Biele, J., Bogatchev, A., Makurin, A., and
  Matrossov, S. (2011).
\newblock Hopper concepts for small body landers.
\newblock \emph{Advances in Space Research} 47, 428--439
\bibAnnoteFile{ulamec2011hopper}

\bibitem[{Wakabayashi et~al.(2009)Wakabayashi, Sato, and
  Nishida}]{wakabayashi2009design}
Wakabayashi, S., Sato, H., and Nishida, S.-I. (2009).
\newblock Design and mobility evaluation of tracked lunar vehicle.
\newblock \emph{Journal of Terramechanics} 46, 105--114
\bibAnnoteFile{wakabayashi2009design}

\bibitem[{Watanabe et~al.(2017)Watanabe, Tsuda, Yoshikawa, Tanaka, Saiki, and
  Nakazawa}]{watanabe2017hayabusa2}
Watanabe, S.-i., Tsuda, Y., Yoshikawa, M., Tanaka, S., Saiki, T., and Nakazawa,
  S. (2017).
\newblock Hayabusa2 mission overview.
\newblock \emph{Space Science Reviews} 208, 3--16
\bibAnnoteFile{watanabe2017hayabusa2}

\bibitem[{Zhang et~al.(2020)Zhang, Zou, Ma, and Wang}]{zhang2020biologically}
Zhang, C., Zou, W., Ma, L., and Wang, Z. (2020).
\newblock Biologically inspired jumping robots: A comprehensive review.
\newblock \emph{Robotics and Autonomous Systems} 124, 103362
\bibAnnoteFile{zhang2020biologically}

\bibitem[{Zhao et~al.(2013{\natexlab{a}})Zhao, Feng, Chu, and
  Ma}]{zhao2013advanced}
Zhao, J., Feng, Z., Chu, F., and Ma, N. (2013{\natexlab{a}}).
\newblock \emph{Advanced theory of constraint and motion analysis for robot
  mechanisms} (Academic Press)
\bibAnnoteFile{zhao2013advanced}

\bibitem[{Zhao et~al.(2013{\natexlab{b}})Zhao, Xu, Gao, Xi, Cintron, Mutka
  et~al.}]{zhao2013msu}
Zhao, J., Xu, J., Gao, B., Xi, N., Cintron, F.~J., Mutka, M.~W., et~al.
  (2013{\natexlab{b}}).
\newblock Msu jumper: A single-motor-actuated miniature steerable jumping
  robot.
\newblock \emph{IEEE Transactions on Robotics} 29, 602--614
\bibAnnoteFile{zhao2013msu}

\end{thebibliography}

\section*{Appendix: The n-sided Sarrus linkage}

\label{appendixA:SCT}

Screws defined by screw theory are used to express the instantaneous velocity and forces acting on a rigid body. Therefore the allowed free motions of a rigid body by its constraints can be represented by screws. A line vector $\bm{S}$ and a couple $\bm{S}_0$ can be combined to form a screw $\bm{\mathbb{S}}$, a dual vector in the so called Pl\"{u}cker homogeneous coordinates consisting of six elements. A screw is represented by four factors in Eq.~\ref{eq_screwb} namely position $\bm{r}$, direction axis and magnitude of the screw $\bm{S}$, and pitch\footnote{$h=\frac{\bm{S}\cdot\bm{S}_0}{\bm{S}\cdot\bm{S}}$} of the screw $h$. For the special case when $\bm{S}\neq0$, $\bm{S}\cdot \bm{S}_0 = 0$, and $h = 0$ ($\bm{S}$ and $\bm{S}_0$ are orthogonal) the screw expresses the Pl\"{u}cker homogeneous coordinates of a line; for $\bm{S}\neq0$, $h = \infty$, $\bm{\mathbb{S}}$ represents a couple. These two special cases are relevant for following descriptions. The relationship between the motions and the constraints of a screw is obtained from reciprocal screw theory and will be discussed later in this section.

\begin{equation}
\bm{\mathbb{S}}=
\begin{bmatrix}
\bm{S}& \bm{S}_0
\end{bmatrix}^T
=
\begin{bmatrix}
\bm{S}&  \bm{r} \times \bm{S}+h \bm{S}
\end{bmatrix}^T
\label{eq_screwb}
\end{equation}

\noindent In the analysis at hand, the degree of freedom of the terminal connector of the $i-$th kinematic chains is analysed first. Figure~\ref{fig_screwsgeom} shows the the schematics of the conceptual mechanism. For geometric convenience, the unit vector $\bm{e_i}$ denotes the direction of the $\pi_i$ plane, while the unit vector $\bm{e_C}$ represents the direction of the intersecting line among planes that can be computed with Eq.~\ref{eq_centrall}. Note that any other nonparallel two planes  may also be used to define $\bm{e_C}$.

\begin{equation}
\bm{e_C}=\frac{\bm{e_1}\times \bm{e_2}}{\left|\bm{e_1}\times \bm{e_2}\right|}
\label{eq_centrall}
\end{equation}

\noindent The screws of the first kinematic chain are obtained with the cartesian coordinates in three-dimensional physical space of the revolute joints $\bm{r_{A_1}}$, $\bm{r_{B_1}}$, and $\bm{r_{C_1}}$ at the fixed coordinate frame. It is important to bear in mind that the screw matrix denotes a six dimensional Euclidean space.

\begin{align*}
\bm{\mathbb{S}}_{A_1}& =
\begin{bmatrix}
\bm{e_1}& \bm{r_{A_1}} \times \bm{e_1} 
\end{bmatrix}^T \\
\bm{\mathbb{S}}_{B_1}&=
\begin{bmatrix}
\bm{e_1}& \bm{r_{B_1}} \times \bm{e_1} 
\end{bmatrix}^T \\
\bm{\mathbb{S}}_{C_1}&=
\begin{bmatrix}
\bm{e_1}& \bm{r_{C_1}} \times \bm{e_1} 
\end{bmatrix}^T
\end{align*}

\noindent The screw of the first kinematic chain can be expressed as the assembly of these three screws of the revolute joints as in Eq.~\ref{eq_kinemch1}. The screw of the $i-$th kinematic chain can be expressed similarly to Eq.~\ref{eq_kinemch1}.

\begin{equation}
\bm{\mathbb{S}}_{A_1B_1C_1}=
\begin{bmatrix}
\bm{\mathbb{S}}_{A_1}&\bm{\mathbb{S}}_{B_1}&\bm{\mathbb{S}}_{C_1}
\end{bmatrix}
\label{eq_kinemch1}
\end{equation}

\noindent The constraints of the $i-$th kinematic chain can be obtained with reciprocal screw theory as mentioned before. Since the physical meaning of the reciprocal product of two screws, \mbox{$\bm{\mathbb{S}}_1\circ \bm{\mathbb{S}}_2=\bm{S}_1\cdot\bm{S}_{0,2} + \bm{S}_2\cdot\bm{S}_{0,1}$}, is the instantaneous work of the force to the motion of the body, the reciprocal product of a screw and its constraint screw must be zero. In other words, no work is produced by a pair of reciprocal screws. Furthermore, the constraint matrix of a fully restricted rigid body is expressed in a six-dimensional Euclidean space, just as the base space of the screw. The reciprocal product $\bm{\mathbb{S}}_{A_1B_1C_1} \circ \bm{\mathbb{S}}_{A_1B_1C_1}^r =0$, uses $\bm{\mathbb{S}}_{A_1B_1C_1}$ denoting the motion of the $i-$th kinematic chain, and $\bm{\mathbb{S}}_{A_1B_1C_1}^r$ is the constraint of the mechanical system acting on it. The reciprocal screw of the $i-$th kinematic chain is given in Eq.~\ref{eq_rkinemch1}

\begin{equation}
\bm{\mathbb{S}}_{A_iB_iC_i}^r = 
\begin{bmatrix}
\bm{\mathbb{S}}_{i}^{r_1} & \bm{\mathbb{S}}_{i}^{r_2} & \bm{\mathbb{S}}_{i}^{r_3}
\end{bmatrix}
\label{eq_rkinemch1}
\end{equation}

\noindent with 

\begin{align*}
\bm{\mathbb{S}}_{i}^{r_1} =&
\begin{bmatrix}
\bm{e_i}& \bm{r_{C_i}} \times \bm{e_i}
\end{bmatrix}^T\\
\bm{\mathbb{S}}_{i}^{r_2}=&
\begin{bmatrix}
0_{1\times3}& \bm{e_C}
\end{bmatrix}^T\\
\bm{\mathbb{S}}_{i}^{r_3}=&
\begin{bmatrix}
0_{1\times3}& \bm{e_C} \times \bm{e_i}
\end{bmatrix}^T
\end{align*}

\begin{figure} 
\centerline{\includegraphics[width=3.0in]{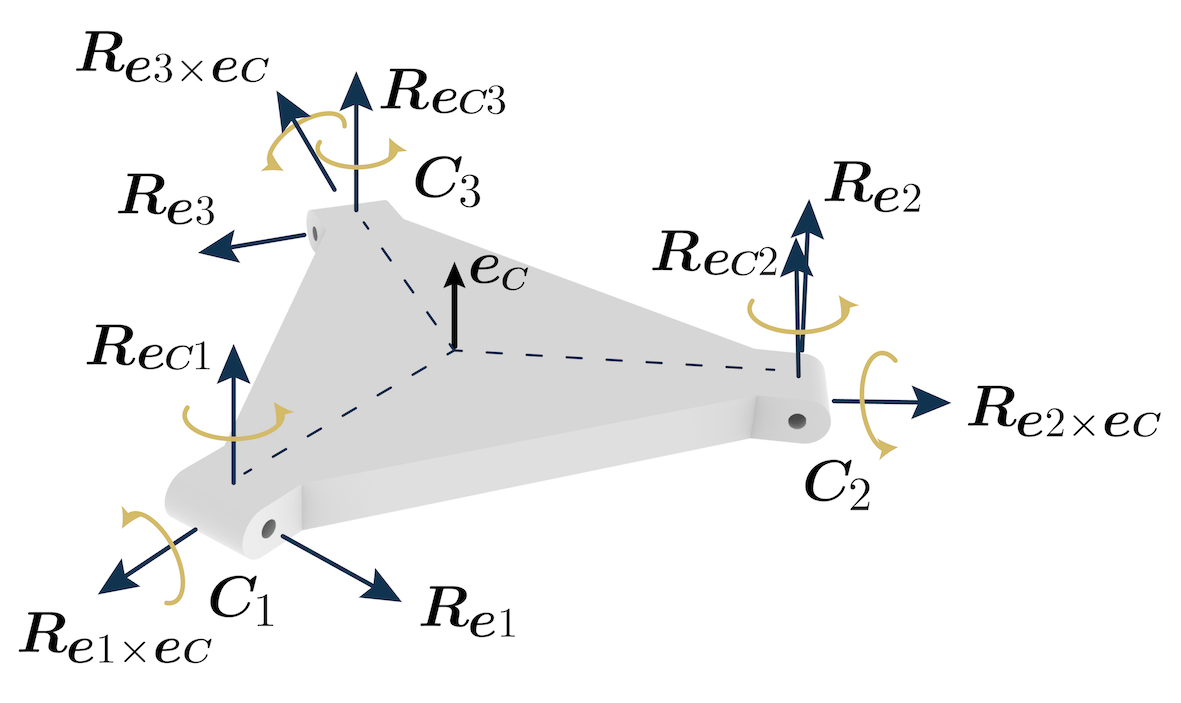}}
\caption{Constraints exerted on the upper platform by the $n$ kinematic chains.}
\label{fig_contrup}
\end{figure}

\noindent By comparing the form of these three constraint screws with Eq.~\ref{eq_screwb}, note that the pitch of $\bm{\mathbb{S}}_{i}^{r_1}$ is zero, which implies a force vector passing through the coordinate $\bm{r_{C_i}}$. Similarly, $\bm{\mathbb{S}}_{i}^{r_2}$ and $\bm{\mathbb{S}}_{i}^{r_3}$ denote torques about their respective axes. In this way, the mechanical system exerts three constraints to the upper platform through this kinematic chain.  
If there exists a reciprocal screw that is reciprocal to all the screws, the reciprocal screw is defined as a common constraint of the mechanism\footnote{Theorem 3.1 in~\cite{huang2013mobility}: a common constraint exists if and only if every limb constraint of the system can provide an identical constraint screw acting on the moving platform, and if these identical screws are constraint forces, they must be coaxial; if these identical screws are couples, they must be parallel.}. Thus, all $n$ kinematic chains exert on the upper platform the common screw $\bm{\mathbb{S}}_{i}^{r_2}$. This represents an over-constraint of a torque about $\bm{e_C}$, see Fig.~\ref{fig_contrup}. The union of constraints of the kinematic chain system on the upper platform can be concatenated as in Eq.~\ref{fig_screwcplat}. Note that the resulting screw, contains only linearly independent screws, i.e. a $i-$th screw represents a linear combination of linearly independent screws.

\begin{equation}
\bm{\mathbb{S}}_{C_1C_2\cdots C_n}^r=
\begin{bmatrix}
\bm{\mathbb{S}}_{1}^{r_1} & \bm{\mathbb{S}}_{1}^{r_2} & \bm{\mathbb{S}}_{1}^{r_3} & \bm{\mathbb{S}}_{2}^{r_1} & \bm{\mathbb{S}}_{2}^{r_3}
\end{bmatrix}
\label{fig_screwcplat}
\end{equation}

\noindent Just as the first reciprocal screw represents dynamic constraints, the second-time reciprocal screw, i.e. the reciprocal screw of a constraint, indicates motion. Consequently the free motion of the upper platform, as well as its constraints are tightly connected by reciprocities. {\textcolor{black}{As the constraints of the upper platform are known,}} Eq.~\ref{fig_screwcplat}, its free motions can be equivalently found through reciprocal screw theory. {\textcolor{black}{The screw in Eq.~\ref{fig_movscrew} is obtained from solving $\bm{\mathbb{S}}_{C_1C_2\cdots C_n}^r \circ \bm{\mathbb{S}}_{C_1C_2\cdots C_n}^P =0$ for $\bm{\mathbb{S}}_{C_1C_2\cdots C_n}^P$,}} which denotes a translation along the vector $\bm{e_C}$ and no rotation freedom in any axis; the same characteristics observed in the classical Sarrus mechanism with two dyads as discussed before. This result implies that the translational mobility remains unchanged independently on the number of kinematic chains added to the mechanism system, as long as there is at least a pair of chains laying in nonparallel $\pi$ planes. Moreover, since all three revolute joints are parallel to the plane normal of the kinematic chain at any configuration, i.e. the axis directions of the three kinematic pairs are always parallel to its characteristic $\bm{e_i}$, the screw expressions are unchangeable, which in turn yields unchangeable common constraints and mobility. Therefore $\bm{\mathbb{S}}_{C_1C_2\cdots C_n}^P$ fully characterises the DOF of the upper platform at any instant. The number of DOFs of the upper platform is simply Rank$\left(\bm{\mathbb{S}}_{C_1C_2\cdots C_i}^P\right)=1$ and the type and direction is expressed by the corresponding screw.

\begin{equation}
\bm{\mathbb{S}}_{C_1C_2\cdots C_n}^P= 
\begin{bmatrix}
0_{1\times3}&\bm{e_C}
\end{bmatrix}^T
\label{fig_movscrew}
\end{equation}

\noindent It is known that the general mobility of a mechanism, specially in parallel mechanisms, sometimes gives no real meanings for both the DOF of the end effector and the actuations required to drive the end effector,~\cite{zhao2013advanced}. Consequently, determining the number of actuations needed to uniquely control the end effector is a complementary analysis to the DOF of the end effector. For the sake of example, {\textcolor{black}{if an actuator were added to cancel the movement of the kinematic pair $B_1$, then the kinematic chain would produce only two screws}}, instead of three as in the other $n$ kinematic chains. The reciprocal screw of this actuated kinematic chain will add a new pure force vector along $\bm{e_C} \times \bm{e_i}$, i.e. a new linearly independent screw. To understand the implications of this result consider the following. Because the Pl\"{u}cker coordinates of a screw system have six components as discussed before, the maximum number of linearly independent screws in a spatial mechanism is always six. A theorem in screw theory states that the summation of the ranks of the screw system and its reciprocal screw is six. {\textcolor{black}{This is acknowledged in the current analysis as}} the three constraints obtained in Eq.~\ref{eq_rkinemch1} are a subspace of the basis Eq.~\ref{eq_kinemch1}, and equivalently for Eq.~\ref{fig_screwcplat} of rank five yields one DOF. The above implies that with the new independent screw constraint in the actuated kinematic chain will increase the rank of the union of constraints of the kinematic chain system on the upper platform (Eq.~\ref{fig_screwcplat}), from five to six. In such case the reciprocal equation has a unique zero solution, which means that the system does not move under the assumed single actuation on $B_1$. {\textcolor{black}{In conclusion from the arguments above, only one independent actuation is required to uniquely control the upper platform of the analysed mechanism consisting of $n-$kinematic chains.}}

\end{document}